\documentclass[letterpaper]{article}
\usepackage{aaai2026}  
\usepackage{times}  
\usepackage{helvet}  
\usepackage{courier}  
\usepackage[hyphens]{url}  
\usepackage{amsmath}
\usepackage{booktabs}
\usepackage{graphicx} 
\usepackage{natbib}  
\usepackage{epigraph}
\usepackage{framed}
\usepackage{placeins}

\usepackage{caption} 
\usepackage{subcaption}
\usepackage{algorithm}
\usepackage{algorithmic}
\usepackage{booktabs}
\usepackage{amssymb}
\usepackage{newfloat}
\usepackage{listings}
\DeclareCaptionStyle{ruled}{labelfont=normalfont,labelsep=colon,strut=off} 
\floatstyle{ruled}
\newfloat{listing}{tb}{lst}{}
\floatname{listing}{Listing}
\title{Talk, Snap, Complain: Validation-Aware Multimodal Expert Framework for Fine-Grained Customer Grievances}
\title{Talk, Snap, Complain: Validation-Aware Multimodal Expert Framework for Fine-Grained Customer Grievances}
\author {
    Rishu Kumar Singh\equalcontrib \textsuperscript{\rm 1},
    Navneet Shreya\equalcontrib \textsuperscript{\rm 2},
    Sarmistha Das\equalcontrib \textsuperscript{\rm 1},
    Apoorva Singh\equalcontrib \textsuperscript{\rm 3},
    Sriparna Saha\textsuperscript{\rm 1}
}
\affiliations {
    \textsuperscript{\rm 1}Indian Institute of Technology Patna, India\\
    \textsuperscript{\rm 2}National Institute of Technology Patna, India\\
    \textsuperscript{\rm 3}Fondazione Bruno Kessler, Italy\\
    \{rishu\_2301ee36, sarmistha\_2221cs21, sriparna\}@iitp.ac.in, navneets.ug23.cs@nitp.ac.in, apoorva0816@gmail.com
}

\usepackage{bibentry}

\begin{document}

\maketitle

\begin{abstract}
Existing approaches to complaint analysis largely rely on unimodal, short-form content such as tweets or product reviews. This work advances the field by leveraging multimodal, multi-turn customer support dialogues—where users often share both textual complaints and visual evidence (e.g., screenshots, product photos)—to enable fine-grained classification of complaint aspects and severity. We introduce \textit{VALOR}, a Validation-Aware Learner with Expert Routing, tailored for this multimodal setting. It employs a multi-expert reasoning setup using large-scale generative models with Chain-of-Thought (CoT) prompting for nuanced decision-making. To ensure coherence between modalities, a semantic alignment score is computed and integrated into the final classification through a meta-fusion strategy. In alignment with the United Nations Sustainable Development Goals (UN SDGs), the proposed framework supports SDG 9 (Industry, Innovation and Infrastructure) by advancing AI-driven tools for robust, scalable, and context-aware service infrastructure. Further, by enabling structured analysis of complaint narratives and visual context, it contributes to SDG 12 (Responsible Consumption and Production) by promoting more responsive product design and improved accountability in consumer services. We evaluate \textit{VALOR} on a curated multimodal complaint dataset annotated with fine-grained aspect and severity labels, showing that it consistently outperforms baseline models, especially in complex complaint scenarios where information is distributed across text and images. This study underscores the value of multimodal interaction and expert validation in practical complaint understanding systems. Resources related to data and codes are available here: https://github.com/sarmistha-D/VALOR
\end{abstract}

\section{Introduction}

\epigraph{Your most unhappy customers are your greatest source of learning.}{\textit{Bill Gates}}
Understanding customer complaints is crucial for improving user experience, service reliability, and product quality, objectives that directly align with the United Nations Sustainable Development Goals (SDGs), particularly SDG 9 (Industry, Innovation, and Infrastructure) and SDG 12 (Responsible Consumption and Production). While prior work has advanced complaint analysis, most efforts focus on short-form, unimodal inputs such as tweets \cite{DBLP:conf/acl/Preotiuc-Pietro19a}. These formats often lack the evolving context and emotional nuance found in real-world customer support scenarios. Moreover, modern users increasingly supplement textual complaints with visual content, such as screenshots of broken interfaces or damaged items shared across platforms \cite{DBLP:conf/aaai/SinghDS022, singh2023your}. However, current research rarely addresses this multimodal and conversational nature of complaints, which introduces unique challenges in cross-modal alignment, contextual reasoning, and fine-grained classification.\\
\begin{figure}[ht]
\centering
\includegraphics[width=6.2cm]
{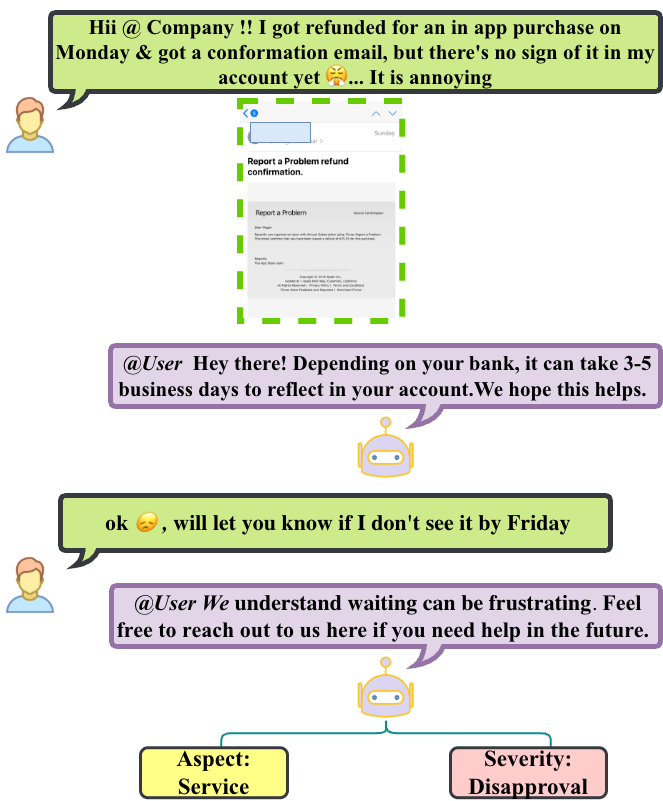} 
\caption{A conversation snippet from the \textit{CIViL} dataset. Labels indicate the aspect-severity pairs identified from the conversation.}
\label{theoretical}
\end{figure}
Traditional approaches like aspect-based sentiment analysis primarily assign sentiment polarities (positive, negative, neutral) to specific aspects in isolated reviews. While helpful in gauging user opinion, these methods fall short when complaints evolve over multi-turn dialogues or involve multimodal evidence. Sentiment alone often lacks the specificity needed for actionable insights. For example, a user might report a late delivery and share a photo of a damaged product, each pointing to different aspects (logistics and packaging) and severity levels. Without disentangling such layered signals, automated systems fail to produce serviceable insights for resolution teams.

We redefine complaint analysis as a fine-grained multimodal classification task over multi-turn dialogues, jointly modeling conversational flow and aligned images to identify aspect categories and severity levels with contextual accuracy. These structured outputs enable downstream applications such as intent-based ticket routing, escalation prediction, and service analytics. Figure~\ref{theoretical} presents an example of such a use case.

Building on this formulation, we propose \textit{CIViL}: Customer Interactions with Visual and Linguistic signals, a curated dataset of multi-turn dialogues annotated with aspect and severity annotations and paired with topically aligned images sourced from social media websites. Motivated by the effectiveness of Large Language Models integrated with Mixture-of-Experts (MoE) architectures, we adopt a modular learning strategy that balances expert specialization with shared representation learning. To this end, we propose \textbf{Validation-Aware Learner with Expert Routing (\textit{VALOR})}, a multimodal framework that transforms complaint interactions into structured outputs using a robust MoE architecture enhanced with semantic alignment and Chain-of-Thought (CoT) reasoning. To assess the quality of modality integration and expert behavior, we incorporate a validation module that applies a three-part metric system, capturing \textit{alignment} across modalities, 
\textit{dominance} when one modality is more informative, and \textit{complementarity} when both modalities contribute distinct yet coherent information. This work lays the foundation for real-world multimodal complaint understanding in dialogue systems by transforming unstructured complaints into structured, fine-grained insights; providing annotated resources, scalable methodologies, and actionable feedback to enhance service responsiveness and consumer engagement.\\
\textbf{Research Objectives: }Following are the research objectives of the current study:\\
(1) To investigate how textual and visual signals enhance fine-grained complaint analysis, specifically aspect and severity detection in multi-turn customer support dialogues.\\
(2) To design and evaluate a two-phase MoE framework, \textit{VALOR}, that combines cross-modal fusion, semantic alignment scoring, and CoT-based expert reasoning to produce accurate predictions from multimodal complaint data.\\
(3) To develop a domain-specific multimodal customer-support dialogue dataset, enabling systematic benchmarking and advancing research in multimodal dialogue-grounded grievance analysis.\\
\textbf{Contributions: }The primary contributions are as follows:\\
(1) We define and investigate the task of fine-grained multimodal complaint understanding within multi-turn dialogues, with a specific emphasis on identifying aspect categories (ACD) and assessing severity levels (SD).\\
(2) We introduce \textit{CIViL}, a benchmark multimodal dataset of customer-support dialogues, built by extending a subset of the Kaggle customer support corpus with fine-grained aspect and severity annotations, and enriched with topically aligned images. This resource aims to advance research in multimodal conversational complaint understanding.\\
(3) We propose \textit{VALOR}, a Validation-Aware Learner with Expert Routing tailored for robust identification of real-world multimodal complaints in conversational settings.\\
(4) The proposed framework sets a strong benchmark for fine-grained multimodal complaint analysis  in dialogue-based contexts, consistently outperforming baselines across multiple metrics.

\section{Related Works}
\subsection{Complaint Detection}
Complaint detection from text has become a key area in computational linguistics. Early methods relied on rule-based systems and handcrafted features to identify dissatisfaction \cite{DBLP:journals/eswa/SinghSHJ21}. With the rise of deep learning, particularly transformer models like BERT \cite{DBLP:conf/naacl/DevlinCLT19} semantic and contextual modeling improved significantly. Recent multitask approaches further enhance generalization by incorporating related cues such as sentiment, emotion, and sarcasm \cite{DBLP:journals/cogcom/SinghSHD22, singh2021you, DBLP:conf/ecir/SinghNS22, singh2023peeking}. However, these studies are limited to short, single-turn texts like tweets which lack contextual depth.\\
Multimodal methods have begun addressing this gap by integrating visual and textual inputs \cite{DBLP:conf/aaai/SinghDS022, DBLP:conf/mm/DevanathanSP024, DBLP:journals/tcss/SinghJDJS24}, but mostly rely on static features or simplistic fusion, often missing nuanced cross-modal interactions. Moreover, evaluations are typically conducted on product reviews, which lack the temporal structure and expressiveness of real-time, multi-turn conversations. Without multi-turn context, these models struggle to capture evolving emotions and layered complaint dimensions. In contrast, multi-turn dialogues offer richer signals, users elaborate issues over time, correct themselves, and discuss multiple concerns not evident in isolated posts.
\begin{table*}[!ht]
\renewcommand{\arraystretch}{0.9}
\centering
\scalebox{0.8}{
\begin{tabular}{c|l}
\hline
\textbf{S.No.} & \textbf{Annotation Guidelines} \\ 
\hline
1 & Annotations must be carried out independently, without outside influence. \\
\hline
2 & Aspect categories should be assigned based on the customer's viewpoint. \\
\hline
3 & Each identified aspect must be paired with an appropriate severity level. \\
\hline
4 & Choose the aspect label that most accurately reflects the specific cause of dissatisfaction.\\
\hline
5 & Ambiguous cases should be resolved through discussion among annotators and authors.  \\
\hline
\end{tabular}}
\caption{Annotation guidelines for \textit{CIViL} dataset.}\label{tabanno}
\end{table*}
\subsection{Fine-grained Complaint Analysis}
Recent advances in complaint analysis have focused on fine-grained tasks like severity classification \cite{DBLP:conf/naacl/JinA21, DBLP:journals/tcss/SinghBS24}, typically using transformer-based models to estimate emotional intensity in social media posts. However, the limited context and ambiguity of short-form content often reduce prediction accuracy. To improve granularity, later work explored aspect-based modeling using attention mechanisms \cite{singh2023investigating, singh2023knowing, jain2023abcord}. While effective, these methods underutilize the deeper reasoning and contextual capabilities of large language models—essential for handling the complexity of aspect and severity classification in rich, multi-layered complaint narratives.
\subsection{Mixture-of-Experts}
Mixture-of-Experts (MoE) architectures enhance model performance by dynamically routing inputs to specialized, lightweight sub-models, each trained on distinct regions of the task space \cite{freund1996experiments}. Unlike traditional ensembles that aggregate outputs post-hoc, modern MoE models employ learnable gating mechanisms to activate only relevant experts during inference, enabling efficient and scalable learning \cite{he2021fastmoe, jiang2024med}. Integrated with large language models (LLMs), these methods have shown strong results across various NLP tasks \cite{DBLP:conf/iclr/ShenHZ0LWCZFCVW24, DBLP:conf/emnlp/LiSYJWX23}, making them suitable for fine-grained complaint analysis in dialogues. More recently, MoE frameworks have been extended to multimodal settings, where experts are trained to process specific modalities or modality combinations, allowing for flexible and fine-grained cross-modal reasoning \cite{yu2024mmoe, li2025uni}. Designing such systems requires attention to expert diversity and routing precision, as misallocation can degrade performance due to loss of critical information.
\subsection{Research Gap}
While automated complaint classification has advanced in recent years, most approaches remain limited to unimodal inputs like text-only tweets or reviews, missing the increasingly common use of visual evidence such as screenshots or product images. Despite this shift in user behavior, current research rarely incorporates visual context into the modeling of customer-support grievances. The interaction between text and images, especially in fine-grained tasks like aspect and severity classification remains underexplored. Multimodal complaint scenarios in dialogue settings introduce unique challenges, including modality alignment, ambiguity resolution, and cross-modal reasoning, which conventional LLMs and vision-language models are not explicitly designed to address.
To address these gaps, we construct a multimodal complaint dataset grounded in customer-support dialogues and propose \textit{VALOR}, a validation-aware multi-expert framework that fuses textual and visual cues via cross-attention, semantic alignment, and Chain-of-Thought reasoning for fine-grained complaint understanding.

\section{Customer Interactions with Visual and Linguistic Signals Dataset (\textit{CIViL}) }
For this study, we build upon the publicly available Kaggle Customer Support on Twitter dataset\footnote{https://www.kaggle.com/datasets/thoughtvector/customer-support-on-twitter}, the only publicly available large-scale collection of English-language customer-agent dialogues across domains such as airlines, tech, and retail \cite{hardalov2018towards}. From over 20,000 conversations, we focused on interactions involving Apple Support, comprising 14\% of the dataset and filtered for two-speaker dialogues ranging from 2 to 10 utterances\footnote{An utterance, also referred to as a turn, typically comprises multiple sentences.} to reflect typical support exchanges. A subset of 2,004 conversations was randomly sampled and manually annotated with fine-grained aspect and severity levels.\\
\begin{table}[!ht]
\renewcommand{\arraystretch}{0.8}
\centering
\scalebox{0.8}{
\begin{tabular}{l|l}
\hline
\textbf{Statistic} & \textbf{Count} \\ 
\hline
Total Conversations & 2004 \\
\hline
Total Utterances & 7101\\
\hline
Total Images & 4478\\
\hline
Total Customer Utterances & 3825 \\
\hline
Total Support agent Utterances & 3276\\
\hline
Average User Utterance per conversation & 2.74 \\
\hline
Average Support agent Utterance per conversation & 1.49  \\
\hline
\end{tabular}}
\caption{\textit{CIViL} dataset statistics}\label{tab1}
\end{table}
To enrich the dataset with visual context, we scraped 4,478 relevant images from the same time period from X\footnote{https://x.com} and Reddit\footnote{https://www.reddit.com} websites using a two-phase pipeline. The scraping process utilized the PRAW library for Reddit and Scrapy\footnote{https://scrapy.org/} for X. First, we curated images related to common complaint themes (e.g., broken screens, battery drain, camera quality) from targeted subreddits. Next, we employed a CLIP-based \cite{radford2021learning} semantic matching algorithm to assign topically aligned images to conversations based on textual content. Only high-confidence image–conversation pairs were retained through a multi-stage similarity and validation process. 
\subsection{Annotator Details}
Three annotators independently labeled each dialogue with aspect and severity labels. Disagreements were resolved through collaborative review sessions, which also involved iterative refinements in the annotation guidelines. The team comprised one Ph.D. researcher and two postgraduate scholars, all with prior experience in supervised dataset creation and domain-specific annotation. Their strong command of English, supported by formal education in English-medium institutions, ensured consistency and accuracy in interpreting conversational content.
\subsection{Annotation Phase \& Dataset Analysis}
To ensure consistency and reduce ambiguity, annotators were equipped with comprehensive guidelines (Table~\ref{tabanno}) and a reference set of  50 sample conversations. The annotation process follows established protocols commonly used in aspect-based sentiment analysis \cite{liao2021improved, nazir2020issues}. 
The \textit{CIViL} dataset comprises the following label distributions:\\
(a) Severity levels-Blame (799), Disapproval (486), Accusation (484), and No Explicit Reproach (235).\\
(b) Aspect categories-Software (1,662), Quality (117), Hardware (112), Service (77), Price (23), and Packaging (13).
Detailed dataset statistics are summarized in Table~\ref{tab1}. 

Fleiss’ Kappa scores \cite{fleiss1971measuring} for inter-annotator agreement were 0.68 for aspect categories and 0.75 for severity levels, indicating substantial consistency among annotators \cite{mchugh2012interrater}. As the dataset exclusively comprises complaint-driven customer-support dialogues, it contains no non-complaint instances. Each conversation is labeled with one or more aspect–severity pairs, capturing the distinct issues raised within a single interaction. In the interest of space, additional annotation details and dataset statistics for \textit{CIViL} are provided in the GitHub repository\footnote{https://github.com/sarmistha-D/VALOR}.\\

\section{Proposed Approach}
\begin{figure*}[t]    
\centering
  \includegraphics[width=0.93\textwidth]{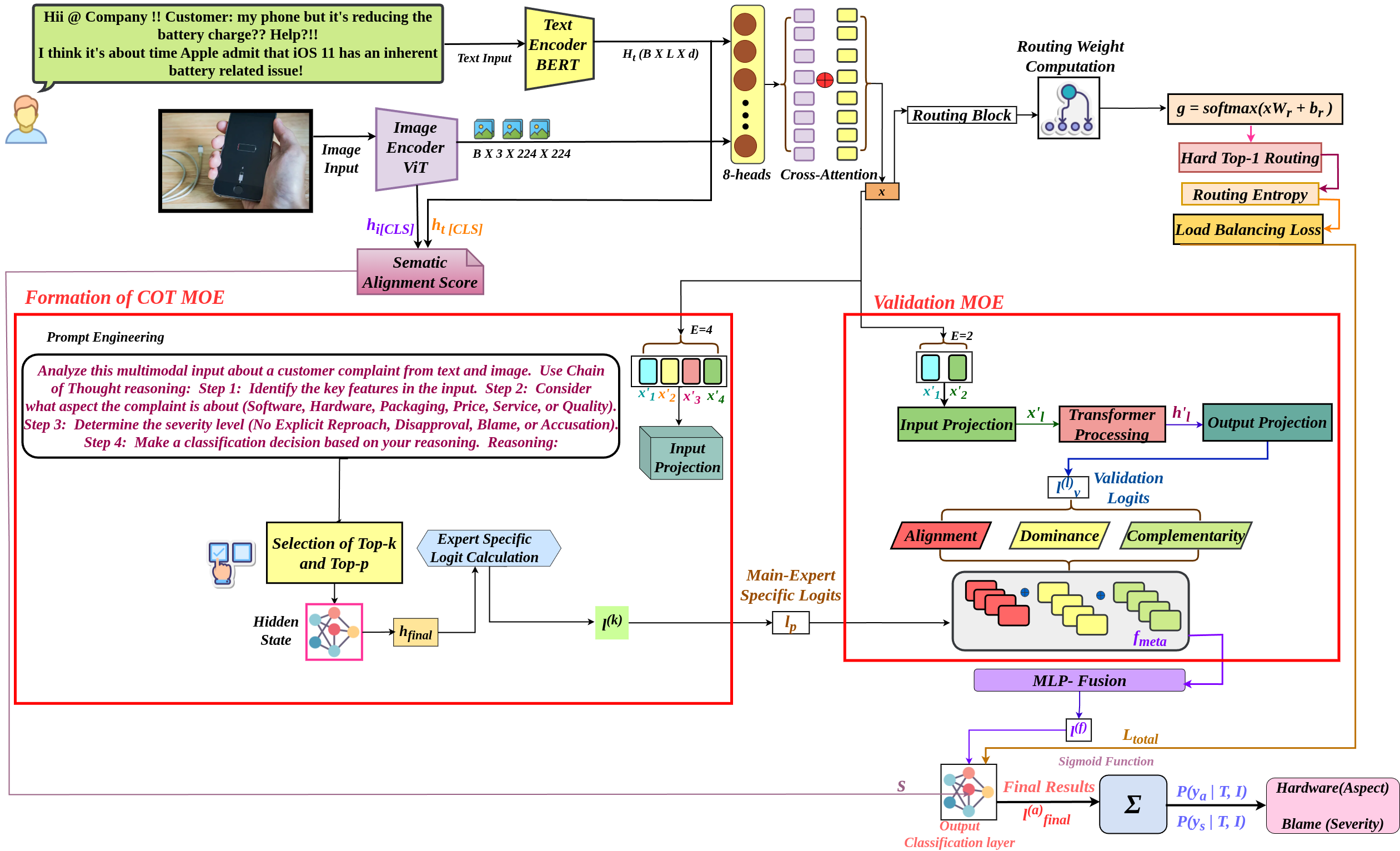}
    \caption{Architectural view of proposed \textit{VALOR} framework}
    \label{archi_Hybrid_MOE}
\end{figure*}

\textbf{Problem Statement: }Given a multimodal input \((T, I)\), where \(T \in \mathbb{R}^{L \times d_T}\) denotes the tokenized textual embedding and \(I \in \mathbb{R}^{3 \times 224 \times 224}\) represents the normalized image tensor, the task is to perform joint classification over two orthogonal dimensions: \textbf{ACD} with \(C_a\) discrete categories (e.g., Software, Hardware, Packaging, etc.) and \textbf{SD} with \(C_s\) ordinal levels (e.g., No Reproach, Disapproval, Blame, Accusation). A unified model \(f_\theta : (T, I) \rightarrow (l_a, l_s)\) is learned, where \(l_a \in \mathbb{R}^{C_a}\) and \(l_s \in \mathbb{R}^{C_s}\) denote the raw logits for aspect and severity, respectively. The predictive distributions are obtained via softmax activation:
\[
P(y_a \mid T, I) = \text{softmax}(l_a), \quad P(y_s \mid T, I) = \text{softmax}(l_s)
\]
enabling efficient end-to-end optimization of the dual-target complaint classification objective within a multimodal learning framework.
We present \textit{VALOR}, a two-step multimodal architecture that combines Chain-of-Thought reasoning with expert validation for fine-grained complaint classification. The framework separates prediction and validation phases, enhancing both accuracy and transparency in multimodal understanding. Figure~\ref{archi_Hybrid_MOE}, outlines the proposed \textit{VALOR} framework.

\begin{enumerate}
    \item \textbf{Phase 1 Prediction (Foundation of COT MoE): }The prediction phase transforms multimodal complaint data into structured outputs through a robust mixture-of-experts framework enhanced with semantic alignment and Chain-of-Thought (CoT) reasoning. Initially, raw text inputs \(T\) are tokenized using the BERT-base-uncased tokenizer (vocabulary size \(V = 30{,}522\)) and truncated to \(L = 512\) tokens, yielding token tensors \(\mathbf{T} \in \mathbb{R}^{\mathcal{B} \times L}\), while image inputs \(I\) are resized to \(224 \times 224\) and passed through a ViT-patch16 ~\cite{DBLP:conf/iclr/DosovitskiyB0WZ21} embedding module to form \(\mathbf{I} \in \mathbb{R}^{\mathcal{B} \times 3 \times 224 \times 224}\) with 196 patches. The text is encoded using a 12-layer, 12-head BERT transformer (hidden size \(d = 768\)) to produce contextual embeddings \(\mathbf{H}_t \in \mathbb{R}^{\mathcal{B} \times L \times d}\) and a [CLS] token \(\mathbf{h}_t \in \mathbb{R}^{\mathcal{B} \times d}\), while the image passes through a ViT-base encoder yielding patch embeddings \(\mathbf{H}_i \in \mathbb{R}^{\mathcal{B} \times 196 \times d}\) and CLS vector \(\mathbf{h}_i \in \mathbb{R}^{\mathcal{B} \times d}\). These representations are fused using cross-modal multi-head attention with \(H = 8\) heads, where for each head \(h\), queries \(\mathbf{Q}_h = \mathbf{H}_t \mathbf{W}_q^{(h)}\), keys \(\mathbf{K}_h = \mathbf{H}_i \mathbf{W}_k^{(h)}\), and values \(\mathbf{V}_h = \mathbf{H}_i \mathbf{W}_v^{(h)}\) are computed using projection matrices \(\mathbf{W}_q^{(h)}, \mathbf{W}_k^{(h)}, \mathbf{W}_v^{(h)} \in \mathbb{R}^{d \times d/H}\). Attention is computed as \(\text{softmax}(\mathbf{Q}_h \mathbf{K}_h^\top / \sqrt{d/H}) \mathbf{V}_h\), and outputs are concatenated, projected, and passed through residual layers and feed-forward networks, followed by mean pooling to yield the unified multimodal embedding \(\mathbf{x} \in \mathbb{R}^{\mathcal{B} \times d}\). Parallelly, a Semantic Alignment Score (SAS) is computed by projecting \(\mathbf{h}_t\) and \(\mathbf{h}_i\) into a shared 512-dimensional space using two-layer MLPs with GELU activation. The outputs are layer-normalized, concatenated, and passed through another MLP with \(\tanh\) activation to yield the scalar alignment score \(s \in [-1, 1]^{\mathcal{B}}\). The fused embedding \(\mathbf{x}\) is then routed to \(\mathcal{K} = 4\)  Chain-of-Thought experts, each built on the DeepSeek-6.7B model (hidden size \(d_t = 4096\)). Each expert \(k\) transforms the input as \(\mathbf{x}'_k = \mathbf{x} \odot \boldsymbol{\alpha}_k + \boldsymbol{\beta}_k\), with learnable parameters \(\boldsymbol{\alpha}_k, \boldsymbol{\beta}_k \in \mathbb{R}^{d}\), followed by projection \(\mathbf{x}''_k = \mathbf{x}'_k \mathbf{W}_{\text{in}}^{(k)} + \mathbf{b}_{\text{in}}^{(k)}\) and autoregressive reasoning using temperature \(\tau = 0.5\), top-\(k = 30\), and top-\(p = 0.9\) sampling. This generates reasoning tokens and a final hidden state \(\mathbf{h}_{\text{final}} \in \mathbb{R}^{\mathcal{B} \times d_t}\), projected as logits \(\boldsymbol{\ell}^{(k)} = \mathbf{h}_{\text{final}} \mathbf{W}_{\text{out}}^{(k)} + \mathbf{b}_{\text{out}}^{(k)} \in \mathbb{R}^{\mathcal{B} \times \mathcal{C}}\), where \(\mathcal{C} \in \{\mathcal{C}_a, \mathcal{C}_s\}\) denotes class counts for aspect or severity prediction. Expert routing is handled by a learned gating function \(\mathbf{g} = \text{softmax}(\mathbf{x} \mathbf{W}_r + \mathbf{b}_r) \in \mathbb{R}^{\mathcal{B} \times \mathcal{K}}\), producing soft probabilities \(g_{b,k}\) for expert relevance. Hard top-1 selection yields expert index \(k^*_b = \arg\max_k g_{b,k}\), resulting in routing matrix \(\mathbf{R} \in \{0,1\}^{\mathcal{B} \times \mathcal{K}}\). To prevent expert collapse and encourage load balancing, the routing entropy \(H(\mathbf{g}_b) = -\sum_{k=1}^{\mathcal{K}} g_{b,k} \log g_{b,k}\) is computed, and a regularization loss is introduced: \(L_{\text{lb}} = \sum_{k=1}^{\mathcal{K}} \left( \frac{1}{\mathcal{K}} - \frac{1}{\mathcal{B}} \sum_{b=1}^{\mathcal{B}} g_{b,k} \right)^2\).

    \item \textbf{Phase 2 (Validation MoE): }The validation phase introduces robust secondary reasoning using \(\mathcal{L}_v = 2\) specialized validation experts and a multi-perspective evaluation pipeline to enhance prediction confidence and interpretability. Each validation expert is instantiated as a DeepSeek transformer stack with 32 layers and hidden dimension \(d_t = 4096\), receiving the joint multimodal embedding \(\mathbf{x} \in \mathbb{R}^{\mathcal{B} \times d}\), which is first linearly projected via \(\mathbf{W}_{v,\text{in}}^{(l)} \in \mathbb{R}^{d \times d_t}\) and bias \(\mathbf{b}_{v,\text{in}}^{(l)}\), yielding \(\mathbf{x}'_l\). This is processed by the transformer block to obtain \(\mathbf{h}_l\), then projected through \(\mathbf{W}_{v,\text{out}}^{(l)}\) and passed through a two-layer MLP with ReLU activations to generate logits \(\boldsymbol{\ell}_v^{(l)} \in \mathbb{R}^{\mathcal{B} \times \mathcal{C}}\). To analyze expert behavior, a threefold metric system is employed: (i) \textit{Alingment} evaluates cosine similarity between logits of experts \(l\) and \(m\), yielding \(\text{Alingment}_{l,m} = \frac{\langle \boldsymbol{\ell}_v^l, \boldsymbol{\ell}_v^m \rangle}{\|\boldsymbol{\ell}_v^l\| \cdot \|\boldsymbol{\ell}_v^m\|}\), with mean score \(R_{\text{avg}}\); (ii) \textit{dominance} quantifies predictive alignment between MoE outputs \(\boldsymbol{\ell}_p\) and validation logits \(\boldsymbol{\ell}_v\) via correlation: \(\text{dominance}^{(a)} = \frac{\text{Cov}(\boldsymbol{\ell}_p^{(a)}, \boldsymbol{\ell}_v^{(a)})}{\sqrt{\text{Var}(\boldsymbol{\ell}_p^{(a)}) \cdot \text{Var}(\boldsymbol{\ell}_v^{(a)})}}\); and (iii) \textit{complementarity}, a diversity measure, is captured via entropy over softmax-normalized logits: \(\text{complementarity}^{(l)} = -\sum_{c=1}^{\mathcal{C}} p_v^{(l,c)} \log p_v^{(l,c)}\), with \(p_v^{(l,c)} = \text{softmax}(\boldsymbol{\ell}_v^{(l)})_c\) and average \(U_{\text{avg}}\). A meta-fusion network then aggregates predictions through routing-aware combination: \(\boldsymbol{\ell}_p = \sum_k \mathbf{R}_{\cdot,k} \cdot \boldsymbol{\ell}^{(k)}\) and \(\boldsymbol{\ell}_v = \sum_l g_{v,l} \cdot \boldsymbol{\ell}_v^l\), where \(g_{v,l}\) are soft routing weights. The combined feature vector \(\mathbf{f}_{\text{meta}} = [\boldsymbol{\ell}_p; \boldsymbol{\ell}_v; s; \bar{H}; R_{\text{avg}}; \text{dominance}; U_{\text{avg}}] \in \mathbb{R}^{\mathcal{B} \times \mathcal{M}}\) (with \(\mathcal{M} = 2\mathcal{C}_a + 5\)) is passed through a 3-layer MLP with hidden sizes \((768, 384, \mathcal{C}_a)\), ReLU activations, and dropout rate 0.1 to generate fused logits \(\boldsymbol{\ell}_f\). These are adjusted by SAS-based alignment: \(\boldsymbol{\ell}_{\text{final}} = \boldsymbol{\ell}_f + \lambda_s \cdot s \cdot \mathbf{1}_{\mathcal{C}_a}\), with \(\lambda_s = 0.1\), and final predictions computed as \(P(y|\cdot) = \text{softmax}(\boldsymbol{\ell}_{\text{final}})\). The overall training objective integrates aspect and severity classification losses using label-smoothed cross-entropy (\(\epsilon_{\text{ls}} = 0.15\)), validation loss \(L_{\text{val}}\), semantic alignment margin loss \(L_{\text{sas}} = \frac{1}{\mathcal{B}} \sum_b \max(0, \mu - s_b)\) with \(\mu = 0.3\), and metric-driven regularizers: \(L_{\text{Alingment}} = \max(0, R_{\text{avg}} - \tau_R)\), \(L_{\text{dominance}} = \max(0, \tau_S - \text{dominance}^{(a)})\), and \(L_{\text{complementarity}} = \max(0, \tau_U - U_{\text{avg}})\), with thresholds \(\tau_R = 0.3\), \(\tau_S = 0.5\), \(\tau_U = 1.5\). The final total loss is computed as: 
\begin{multline*}
L_{\text{total}} = L_{\text{aspect}} + L_{\text{severity}} + \lambda_{\text{lb}} L_{\text{lb}} + \lambda_{\text{val}} L_{\text{val}} \\
+ \lambda_s L_{\text{sas}} + \lambda_R L_{\text{Alingment}} + \lambda_S L_{\text{dominance}} + \lambda_U L_{\text{complementarity}},
\end{multline*}
The final predictions for both aspect and severity are computed as $P(y_a \mid T, I) = \text{softmax}(l_{final}^{(a)})$ and $P(y_s \mid T, I) = \text{softmax}(l_{final}^{(s)})$,  respectively.
\end{enumerate}

\section{Experiments and Results}
This section describes the evaluation setup used to ensure a fair and rigorous comparison with strong state-of-the-art baselines. Our analysis is driven by three research questions: \textbf{RQ1: }How does \textit{VALOR} perform relative to state-of-the-art models? \\ \textbf{RQ2: }What is the individual contribution of each architectural component in \textit{VALOR}?\\  \textbf{RQ3: }Which expert configuration delivers optimal task performance?
\subsection{Evaluation Protocol}
We evaluate the proposed \textit{VALOR} framework on the \textit{CIViL} dataset. The dataset is split into 70\% training, 10\% validation, and 20\% testing to ensure statistical robustness and generalization fidelity. All experiments are conducted on an NVIDIA RTX 3090 GPU, using the AdamW optimizer\footnote{https://keras.io/api/optimizers/adamw/} with an initial learning rate of $2 \times 10^{-5}$, linear warm-up, and cosine decay scheduling. Optimization employs \textit{binary cross-entropy loss} for multi-label classification, with a batch size of 16, dropout rate of 0.5, and gradient clipping (max norm = 1.0) to stabilize training. Models are trained for up to 20 epochs with early stopping (patience = 5) based on validation loss, and a fixed random seed (42) ensures full reproducibility. Evaluation is performed using Accuracy and macro F1-score, computed independently for both ACD and SD tasks to provide a comprehensive assessment of model performance across the two fine-grained complaint dimensions.

\subsection{Baseline Comparison}
To rigorously benchmark the performance of our \textit{VALOR} framework, we evaluate it against a broad suite of state-of-the-art multimodal models across three learning paradigms, \textit{zero-shot}, \textit{few-shot}, and \textit{fully fine-tuned}, to assess adaptability under varying supervision levels. The baselines include leading \textbf{vision-aligned language models} such as DeepSeek-VL, Gemma-3 (9B), Flash Gemini (1.6B), and Paligemma (3B), as well as prominent \textbf{vision-language pretraining architectures} like ImageBind, SMOL-VLM, GIT (300M), FLAVA, ALBEF, UNITER, CLIP ViT-B/32, VisualBERT, and ViLT. Together, these models represent the state-of-the-art in multimodal representation learning. As shown in Table~\ref{tab:baseline_results}, we report and rank their performance after 20 epochs of fine-tuning on the \textit{CIViL} dataset, providing a comprehensive comparison across both aspect and severity prediction tasks.

\begin{table}[h!]
\centering
\small
\setlength{\tabcolsep}{2.5pt}
\renewcommand{\arraystretch}{0.8}
\begin{tabular}{@{}lcccccc@{}}
\toprule
\textbf{Model} & \multicolumn{2}{c}{\textbf{ACD}} & \multicolumn{2}{c}{\textbf{SD}} \\
\midrule
& \textbf{A} & \textbf{F1} & \textbf{A} & \textbf{F1}\\
\hline
DeepSeek-VL & 0.66 & 0.65 & 0.66 & 0.65  \\
Gemma-3 (9B) & 0.69 & 0.66 & 0.65 & 0.66  \\
Flash Gemini (1.6B) & 0.66 & 0.65 & 0.66 & 0.65  \\
ImageBind & 0.66 & 0.65 & 0.64 & 0.63 \\
Paligemma (3B) & 0.65 & 0.66 & 0.65 & 0.64  \\
SMOL-VLM & 0.65 & 0.64 & 0.63 & 0.62  \\
GIT (300M) & 0.65 & 0.64 & 0.63 & 0.62  \\
FLAVA & 0.62 & 0.61 & 0.60 & 0.59  \\
ALBEF & 0.61 & 0.60 & 0.59 & 0.56  \\
UNITER & 0.60 & 0.59 & 0.56 & 0.55  \\
CLIP ViT-B/32 & 0.59 & 0.56 & 0.55 & 0.56 \\
VisualBERT & 0.56 & 0.55 & 0.56 & 0.55  \\
ViLT & 0.55 & 0.56 & 0.55 & 0.54  \\
\bottomrule
\end{tabular}
\caption{Performance comparison of baseline models on the \textit{CIViL} dataset after 20 epochs of fine-tuning. Models are ranked by their overall F1-score. A: Accuracy, F1: macro F1-score}
\label{tab:baseline_results}
\end{table}

The results indicate that larger models such as Gemma-3 (9B) tend to perform better, suggesting that model scale is a significant factor in this complex task. Models specifically designed for vision-language integration, like DeepSeek-VL, also show strong performance. However, there is considerable variance across the board, underscoring the challenges of fine-grained multimodal analysis.

\subsection{Ablation Study}
To quantify the impact of key design choices in the \textit{VALOR} framework, we conduct a focused ablation study across four core components. First, we compare our Chain-of-Thought (CoT) experts with standard MLP and Transformer-based experts. Second, we assess the effect of including the \textit{Validation MoE} module. Third, we evaluate our learnable \textit{Semantic Alignment Score} (SAS) against cosine similarity and alignment-agnostic baselines. Lastly, we vary the \textit{Top-K routing} parameter to analyze the influence of expert sparsity. Results in Table~\ref{tab:ablation_results} provides proposed framework \textit{VALOR} result and detailed insights into the role of each component. 
\begin{table*}[!t]
\centering
\small
\setlength{\tabcolsep}{2.5pt}
\begin{tabular}{@{}lcccccccc@{}}
\toprule
\textbf{Configuration} & \textbf{Expert} & \textbf{Validation MoE} & \textbf{SAS} & \textbf{Top-K} & \textbf{ACD (A)} & \textbf{SD (A)} & \textbf{ACD (F1)} & \textbf{SD (F1)} \\
\midrule
CoT (No Validation, Learnable SAS, Top-2)& cot & False & learnable & 2 & 73.74 & 62.62 & 70.44 & 52.84 \\
CoT (No Validation, Learnable SAS, Top-4)& cot & False & learnable & 4 & 75.14 & 64.64 & 70.46 & 59.47 \\
\textbf{\textit{VALOR}}& \textbf{cot} & \textbf{True} & \textbf{learnable} & \textbf{2} & \textbf{81.94} & \textbf{72.51} & \textbf{76.96} & \textbf{67.91} \\
MLP (No Validation, Learnable SAS, Top-2)& mlp & False & learnable & 2 & 70.43 & 57.35 & 63.82 & 48.55 \\
MLP (No Validation, Learnable SAS, Top-4)& mlp & False & learnable & 4 & 71.97 & 58.98 & 65.04 & 54.45 \\
MLP (Validation, Cosine SAS, Top-2)& mlp & True & cosine & 2 & 68.81 & 65.24 & 61.20 & 57.58 \\
MLP (Validation, Cosine SAS, Top-4)& mlp & True & cosine & 4 & 69.06 & 65.14 & 66.62 & 56.14 \\
MLP (Validation, No SAS, Top-2)& mlp & True & none & 2 & 71.19 & 64.39 & 68.18 & 60.69 \\
Transformer (No Validation, Learnable SAS, Top-2)& transformer & False & learnable & 2 & 77.51 & 67.95 & 70.62 & 61.70 \\
Transformer (No Validation, No SAS, Top-2)& transformer & False & none & 2 & 72.86 & 52.67 & 68.22 & 43.84 \\
Transformer (No Validation, No SAS, Top-4)& transformer & False & none & 4 & 65.51 & 59.48 & 59.79 & 55.17 \\
Transformer (Validation, Learnable SAS, Top-2)& transformer & True & learnable & 2 & 77.08 & 63.98 & 70.24 & 60.24 \\
Transformer (Validation, No SAS, Top-2)& transformer & True & none & 2 & 74.84 & 63.55 & 71.30 & 58.05 \\
\bottomrule
\end{tabular}
\caption{Results for \textit{VALOR} framework and ablation study of its different components. The table shows the performance impact of different expert types, validation strategies, and SAS settings. Best scores are in bold face. A: Accuracy, F1: macro F1-score}
\label{tab:ablation_results}
\end{table*}

\subsection{Results Analysis}
Our experimental results are guided by three research questions (RQs):

\textbf{RQ1: How does \textit{VALOR} perform relative to state-of-the-art models?}
In its full configuration, \textit{VALOR} achieves significant performance gains over all competitive baselines, attaining 81.94\% aspect classification accuracy and 72.51\% severity accuracy, marking absolute improvements of 12.94\% and 6.51\%, respectively, over the strongest contender, \textit{Gemma-3}. These results empirically validate the efficacy of our expert-driven, validation-aware architecture and underscore its superior capacity for multimodal complaint understanding.

\textbf{RQ2: What is the contribution of each component in \textit{VALOR}?}
The ablation analysis substantiates the additive efficacy of each architectural module within \textit{VALOR}. Notably, the incorporation of the \textit{Validation MoE} yields a substantial gain of 8.2\% in aspect accuracy (from 73.74\% to 81.94\%), highlighting its pivotal role in expert quality control. Furthermore, the integration of a learnable \textit{Semantic Alignment Score (SAS)} consistently outperforms static alignment baselines (e.g., cosine similarity or no alignment), affirming the importance of dynamic cross-modal alignment in enhancing representational fidelity and task-specific reasoning.

\textbf{RQ3: Which expert type is most effective for this task?}
\textit{Chain-of-Thought (CoT)} experts exhibit clear superiority over both Transformer and MLP-based counterparts, attributed to their explicit step-by-step reasoning capabilities that are essential for capturing the nuanced semantics of customer complaints. While Transformer experts offer competitive performance due to their expressive capacity, they fall short in interpretability and sequential inference. MLP experts, limited by their architectural simplicity and lack of contextual modeling, yield the weakest results, underscoring the critical advantage of structured reasoning in this domain. \textit{All reported results are statistically significant\footnote{We performed Student's t-test for the test of significance. The results are statistically significant when testing the null hypothesis (p-value $<$ 0.05).} \cite{welch1947generalization}.}

\begin{figure}
    \centering
    \includegraphics[width=0.82\linewidth]{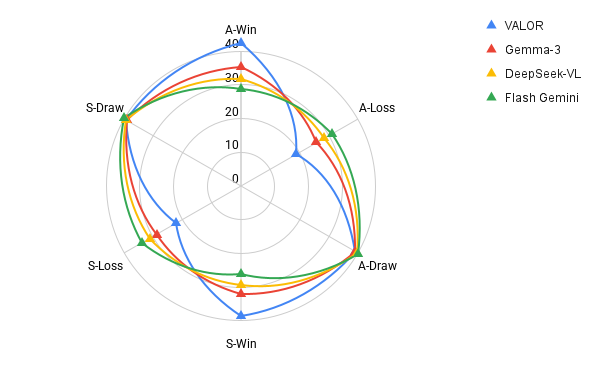}
    \caption{Human Evaluation on win-loss-draw performance criteria between popular baselines against \textit{CIViL}; Here A-Stands for Aspect and S-stands for Severity}
    \label{fig:human}
\end{figure}

\paragraph{Human Evaluation: }To provide a fine-grained and context-aware assessment of \textit{VALOR}'s real-world effectiveness beyond conventional automated metrics, we conducted a rigorous human evaluation (Figure \ref{fig:human}) involving 200 randomly selected test samples from the \textit{CIViL} dataset, encompassing diverse complaint categories and severity levels. Using a win-loss-draw protocol, expert evaluators compared \textit{VALOR}’s predictions against those of state-of-the-art baselines, Gemma-3 (a reasoning-aligned encoder-decoder LLM), DeepSeek-VL, and Flash Gemini, across two core dimensions: aspect identification and severity classification. \textit{VALOR} demonstrated superior judgmental fidelity, achieving the highest win rates at 42.3\% for aspect identification and 38.5\% for severity classification, while simultaneously maintaining the lowest loss rates of 18.7\% and 22.1\%, respectively. These empirical findings underscore \textit{VALOR}’s architectural advantage in integrating validation-aware multimodal experts, Chain-of-Thought-enabled expert modules, and semantic alignment mechanisms, which collectively enhance its interpretability and robustness in nuanced complaint understanding tasks.
\paragraph{Qualitative Analysis: }To gain deeper insight into the behavior of \textit{VALOR} framework, we conducted a focused qualitative analysis of its predictions. Key observations include:\\
(1) Multi-aspect recognition: \textit{VALOR} effectively handles complex complaints involving multiple issues. For example, a user reporting a “battery issue and slow software,” accompanied by an image indicating the battery problem, is accurately classified into distinct aspect–severity pairs: hardware–disapproval and software–accusation.\\
(2) Expert fallback reliability: The validation-aware MoE layer successfully intervenes in low-confidence cases from the primary experts, offering a corrective secondary inference. In the interest of space, additional qualitative analyses and representative examples are provided in the GitHub repository\footnote{https://github.com/sarmistha-D/VALOR}.
\paragraph{Error Analysis: }The error analysis of \textit{VALOR} highlights following key challenges impacting aspect–severity classification in conversational settings:\\ 
(1) \textit{Subjective severity interpretation:} Variability in user tone or emotionally neutral expressions can lead the model to underestimate or misclassify severity levels.\\
(2) \textit{Class imbalance:} Over representation of dominant aspects (e.g., “software”) and underrepresentation of others (e.g., “price”) skew predictions and reduce generalization to low-frequency categories. This distribution, while challenging, mirrors real-world complaint frequency patterns.
\section{Conclusion and Future Work}
This work introduces \textit{CIViL}, a benchmark multimodal dialogue dataset, and \textit{VALOR}, a modular Mixture-of-Experts framework for fine-grained multimodal complaint analysis in customer-support conversations. By integrating semantic alignment, Chain-of-Thought reasoning, and expert validation, \textit{VALOR} produces structured, interaction-aware predictions that consistently outperform strong baselines. Our experiments highlight the effectiveness of dynamic expert routing in disentangling complex, multimodal signals across dialogue turns. Future work will focus on extending the framework to support multilingual scenarios and additional multimodal signals, while incorporating speaker roles and temporal dependencies to enhance its applicability across diverse service contexts and user populations.\\
\section*{Ethics Statement}This work is conducted solely for the research community and is not intended for commercial use. Neither the authors nor the annotators intend to defame any company. Authors and annotators refrained from expressing personal views during the dataset creation process.

\bibliography{main}

\section{Appendix}

\subsection{\textit{CIViL} Dataset details}

\subsubsection{Phase 1: Image Corpus Curation from Reddit}

Given the API limitations of the original data source, we turned to Reddit as a rich source of user-generated visual content related to product issues. We automated the image collection process using a Python script built upon the \textbf{PRAW (Python Reddit API Wrapper)} library.\\
Our strategy was designed to maximize both the quantity and relevance of the collected images.

\subsubsection{Targeting Strategy}

We identified 15 distinct problem categories and compiled a list of associated search keywords for each. The categories covered a wide range of common user complaints:
\begin{itemize}
\item Typing and autocorrect errors
\item Bugs and glitches after software updates
\item Rapid battery drain
\item General dissatisfaction and negative feedback
\item Physical screen damage (cracks, flickering)
\item Performance lag and freezing
\item Camera quality and functionality issues
\item Poor sound or call quality
\item Water damage
\item Charging port and cable problems
\item ``Storage full'' errors
\item Issues with specific third-party applications
\item Connectivity problems (Wi-Fi, Bluetooth, Cellular)
\item Malfunctioning accessories (e.g., AirPods, Apple Watch)
\item Setup and activation errors
\end{itemize}

To capture a broad spectrum of content, we scraped from a diverse set of nine subreddits, including general tech communities (r/gadgets), brand-specific forums (r/apple, r/iphone), and technical support channels (r/techsupport, r/mobilerepair).

\subsubsection{Comprehensive Scraping}

For each subreddit, we queried posts from multiple endpoints: hot, top (with yearly and monthly filters), and new. We also performed explicit keyword searches for each term within our 15 predefined categories to ensure targeted retrieval.

\subsubsection{Quality and Relevance Filtering}

To maintain a high-quality corpus, we applied several filtering criteria to each potential post. The post's title or body had to contain a general keyword (e.g., ``iphone'', ``apple'', ``ios''); it needed a minimum upvote score; the URL had to point directly to an image file; and the downloaded image's resolution had to exceed 50,000 total pixels to avoid low-quality thumbnails.

\subsubsection{Metadata Preservation}

Upon successful validation, we saved each image with a structured filename that embedded crucial metadata\footnote{category\_\_subreddit\_\_score\{score\}\_\_\{term\}\_\_\{post\_id\}.jpg}. This allowed us to retain the image's original context, popularity, and the search category that discovered it.

This process yielded a final corpus of \textbf{4,478} unique, contextually relevant images, which formed the foundation for our mapping phase.
\begin{table*}[!ht]
\centering
\renewcommand{\arraystretch}{1.1} 
\setlength{\tabcolsep}{10pt} 

\begin{tabular}{p{8cm}|c|c|c}
\hline
\textbf{Conversation} & \textbf{Image} & \textbf{Aspect} & \textbf{Severity} \\ \hline

\raggedright
@Company hey my i on my keyboard isn’t working. I just updated my phone to the latest IOS.\\
@Customer We understand your concern. We are working on it and will correct it very soon. Sincere apologies for inconvenience.\\
@Company Release an update soon, and notify me of it. 
& \raisebox{-0.5\height}{\includegraphics[width=2cm]{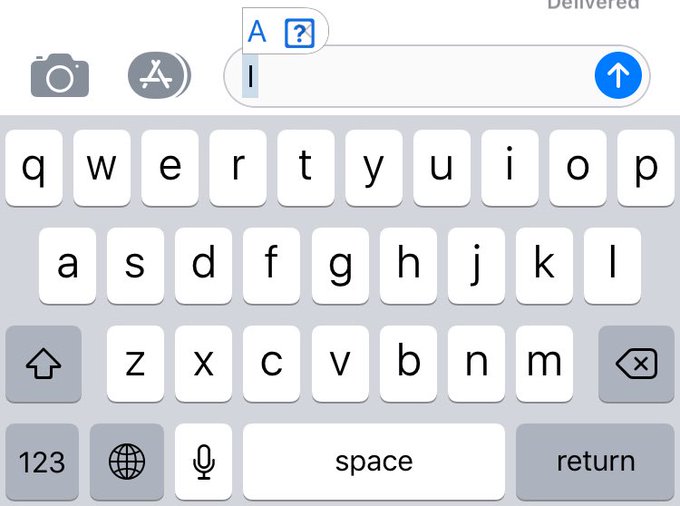}} 
& Software & Disapproval \\ \hline

\raggedright
@Company Why is the new update savaging my battery, 100\% to 1\% in the span of an hour, get your shit together.\\
@Customer We understand your concern. It happens in updates, should stabilize.\\
@Company You released faulty software, and my device is now practically unusable.\\
@Customer Please reach out to us if not resolved in 24 hrs. We are here to help. 
& \raisebox{-0.5\height}{\includegraphics[width=2cm]{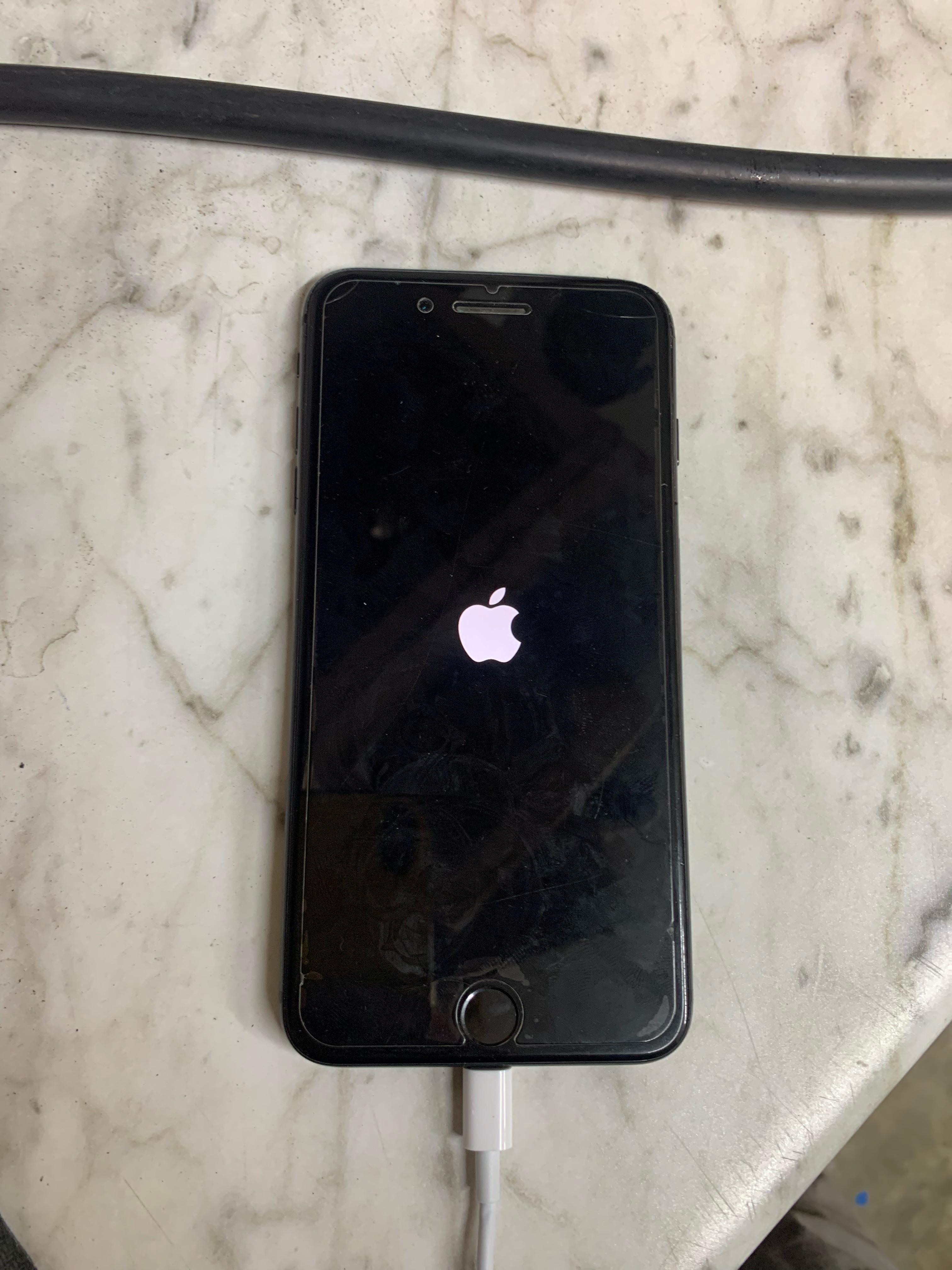}} 
& Software & Blame \\ \hline

\end{tabular}
\caption{Sample conversations from the \textit{CIViL} dataset.}
\label{tableSAMP}
\end{table*}

\subsubsection{Phase 2: Vision-Language Model-Based Assignment}

The core of our contribution lies in the sophisticated method we devised to assign the most relevant image to each conversation thread. We moved beyond simple keyword matching and instead leveraged the semantic understanding of advanced vision-language models.

\subsubsection{Model Roles: CLIP for Matching, BLIP for Analysis}
We employed two distinct models, \textbf{CLIP} and \textbf{BLIP}, for different purposes in our pipeline.

\paragraph{CLIP\footnote{clip-vit-base-patch32} \cite{radford2021learning}: }This model was the cornerstone of our matching algorithm. CLIP's ability to embed both images and text into a shared vector space allowed us to quantitatively measure the semantic similarity between them. We used it to calculate the alignment between an image and three different facets of a conversation: its textual content, its annotated aspect, and its annotated severity.

\paragraph{BLIP\footnote{blip-image-captioning-base} \cite{li2022blip}: }In contrast, we used BLIP primarily for \textbf{supplementary analysis and reporting}. After an image was assigned to a conversation by our CLIP-based algorithm, we used BLIP to generate a descriptive caption for that image. We then calculated a simple word-overlap similarity between this generated caption and the original conversation text. This BLIP-based similarity score was recorded for analysis but was \textbf{not used in the primary decision-making process} for assigning an image.

\subsubsection{Multi-Dimensional Assignment Algorithm}

Our assignment logic is a hierarchical process designed to ensure a high degree of semantic relevance for every match.

\paragraph{Multi-Faceted Embedding: }First, we embedded all components into CLIP's vector space. This included the scraped images, the context-enhanced conversation texts (text combined with its aspect and severity labels), and, crucially, the \textbf{annotation labels themselves}. We generated prompts like ``This complaint is about Hardware.'' to create a target vector for each annotation category.

\paragraph{Three-Pronged Similarity Scoring: }For every (conversation, image) pair in our dataset, we calculated three distinct CLIP cosine similarity scores:

\begin{itemize}
  \item Text-to-Image Similarity ($S_{\text{text}}$)
   \item Aspect-to-Image Similarity ($S_{\text{aspect}}$)
   \item Severity-to-Image Similarity ($S_{\text{severity}}$)
\end{itemize}

\paragraph{Hierarchical Thresholding and Selection: }An image was only assigned to a conversation if it satisfied a rigorous, multi-step validation process. First, an image had to pass all three individual similarity thresholds to be considered a candidate match. For all candidate images that passed this initial filter, we calculated a final combined score using predefined weights:

\begin{equation}
S_{\text{combined}} = (w_{\text{text}} \cdot S_{\text{text}}) + (w_{\text{aspect}} \cdot S_{\text{aspect}}) + (w_{\text{severity}} \cdot S_{\text{severity}})
\end{equation}

We assigned the image with the highest $S_{\text{combined}}$ to the conversation, but only if this score also surpassed a final global threshold. If no image for a given conversation could satisfy this entire chain of criteria, no image was assigned, ensuring that only high-confidence matches were included in the final dataset.
Table \ref{tableSAMP} shows few sample conversations from the \textit{CIViL} dataset.
 

\subsection{\textit{VALOR} Methodology details}
This section provides additional implementation details for the \textit{VALOR} framework, including hyperparameters, architectural specifications, and training configurations. 

\paragraph{Chain-of-Thought Expert Configuration: }The Chain-of-Thought (CoT) experts form the core reasoning component of our framework. Each expert is built upon the DeepSeek-6.7B model and employs structured reasoning to analyze multimodal inputs. The experts are designed to perform step-by-step analysis of customer complaints, enabling interpretable decision-making processes.\\
\textbf{Prompt Engineering Strategy: }Each CoT expert employs a carefully crafted 4-step reasoning prompt designed specifically for multimodal complaint analysis. The prompt structure guides the model through a systematic analysis process, ensuring comprehensive evaluation of both textual and visual information. The prompt template is designed to encourage explicit reasoning while maintaining consistency across different expert instances.

\begin{quote}
\small
\textbf{Chain-of-Thought Prompt Template:}
\begin{framed}
\begin{itshape}
``Analyze this multimodal input about a customer complaint from text and image. 

Use Chain of Thought reasoning:

Step 1: Identify the key features in the input.

Step 2: Consider what aspect the complaint is about (Software, Hardware, Packaging, Price, Service, or Quality).

Step 3: Determine the severity level (No Explicit Reproach, Disapproval, Blame, or Accusation).

Step 4: Make a classification decision based on your reasoning.

Reasoning:''
\end{itshape}
\end{framed}
\end{quote}

This structured approach ensures that each expert follows a consistent reasoning pattern while allowing for expert-specific insights. The prompt is designed to be generic enough to handle various complaint types while specific enough to guide the model toward the target classification tasks.

\paragraph{Generation Parameters and Control: }To ensure high-quality reasoning outputs, each expert employs carefully tuned generation parameters. The temperature parameter is set to $\tau = 0.5$ to balance creativity with consistency, allowing for controlled randomness in the reasoning process while maintaining coherent output. Top-k sampling with $k = 30$ provides vocabulary diversity without overwhelming the model with too many options.

Nucleus sampling with $p = 0.9$ ensures that the model focuses on the most probable tokens while maintaining some flexibility. The maximum token generation is limited to $L_{\text{max}} = 24$ reasoning tokens per sample, which provides sufficient space for detailed reasoning while preventing excessive verbosity.

\paragraph{Expert-Specific Parameterization}

Each CoT expert $k$ employs learnable scale and bias parameters to enable expert-specific input transformation. This parameterization allows each expert to develop specialized processing capabilities while maintaining the overall architectural consistency. The transformation is defined as:

\begin{align}
\mathbf{x}'_k &= \mathbf{x} \odot \boldsymbol{\alpha}_k + \boldsymbol{\beta}_k \\
\boldsymbol{\alpha}_k, \boldsymbol{\beta}_k &\in \mathbb{R}^{d}
\end{align}

where $\boldsymbol{\alpha}_k$ and $\boldsymbol{\beta}_k$ are learnable parameters that enable expert-specific input transformation. This approach allows each expert to adapt its processing to different types of complaints while maintaining the overall framework structure.

\subsubsection{Validation Expert Architecture}

The validation experts serve as a secondary reasoning layer, providing robust verification of the primary predictions. These experts are designed to complement the CoT experts by offering different perspectives on the same multimodal input, enhancing the overall reliability of the system.

\paragraph{Transformer Configuration and Efficiency}

Each validation expert utilizes a 32-layer DeepSeek transformer with a hidden dimension of $d_t = 4096$. To balance computational efficiency with model capacity, we employ a selective fine-tuning strategy where only the last 2 layers are updated during training, while the first 30 layers remain frozen. This approach maintains the rich pretrained knowledge while significantly reducing computational overhead.

The frozen layers preserve the extensive knowledge acquired during pretraining, while the fine-tuned layers adapt to the specific task requirements. This strategy has been empirically shown to provide the best balance between performance and computational efficiency for our multimodal classification task.




\paragraph{Router Configuration and Load Balancing: }The expert router plays a crucial role in ensuring balanced utilization of all experts while preventing expert collapse. The router employs sophisticated mechanisms to distribute inputs appropriately across the expert ensemble, ensuring that each expert contributes meaningfully to the overall system performance.

\paragraph{Load Balancing Mechanisms: }The expert router implements several mechanisms to prevent expert collapse and encourage balanced expert utilization. Noise injection with standard deviation $\sigma = 0.05$ provides regularization during training, preventing the router from becoming overly deterministic. The load balance weight of $\lambda_{\text{lb}} = 0.05$ ensures that the load balancing objective receives appropriate attention during optimization.

The routing strategy employs hard top-1 selection, which ensures that each input is routed to exactly one expert. This deterministic routing strategy simplifies the system while maintaining the benefits of expert specialization. The entropy regularization term $H(\mathbf{g}_b) = -\sum_{k=1}^{\mathcal{K}} g_{b,k} \log g_{b,k}$ encourages the router to maintain uncertainty in its decisions, preventing premature convergence to a single expert.

\subsubsection{Analysis Metrics and Evaluation}

The framework employs a comprehensive set of analysis metrics to evaluate expert behavior and system performance. These metrics provide insights into the internal dynamics of the expert ensemble and help identify potential areas for improvement.

\paragraph{Alignment Analysis: }The alignment analysis measures the similarity between validation experts using cosine similarity:

\begin{align}
\text{Alignment}_{l,m} = \frac{\langle \boldsymbol{\ell}_v^l, \boldsymbol{\ell}_v^m \rangle}{\|\boldsymbol{\ell}_v^l\| \cdot \|\boldsymbol{\ell}_v^m\|}
\end{align}

This metric quantifies the degree of agreement between different validation experts, providing insights into the consistency of the validation process. High alignment scores indicate that the validation experts are converging on similar predictions, while low scores may indicate diverse perspectives or potential issues with the validation process.

\paragraph{Dominance Analysis: }The dominance analysis measures the correlation between the main MoE predictions and validation expert predictions:

\begin{align}
\text{dominance}^{(a)} = \frac{\text{Cov}(\boldsymbol{\ell}_p^{(a)}, \boldsymbol{\ell}_v^{(a)})}{\sqrt{\text{Var}(\boldsymbol{\ell}_p^{(a)}) \cdot \text{Var}(\boldsymbol{\ell}_v^{(a)})}}
\end{align}

This metric quantifies the extent to which the validation experts agree with the primary predictions. High dominance scores indicate strong agreement between the main and validation predictions, while low scores may indicate that the validation experts are providing different perspectives or identifying potential issues with the primary predictions.

\paragraph{Complementarity Analysis: }The complementarity analysis measures the diversity of validation expert predictions using entropy:

\begin{align}
\text{complementarity}^{(l)} = -\sum_{c=1}^{\mathcal{C}} p_v^{(l,c)} \log p_v^{(l,c)}
\end{align}

where $p_v^{(l,c)} = \text{softmax}(\boldsymbol{\ell}_v^{(l)})_c$. This metric quantifies the uncertainty or diversity in the validation expert predictions. High complementarity scores indicate diverse expert opinions, while low scores indicate consensus among the validation experts.

\subsubsection{Training Configuration and Optimization}

The training process employs sophisticated optimization strategies to ensure stable convergence and optimal performance. The configuration balances multiple objectives while maintaining computational efficiency.

\paragraph{Optimizer Configuration:} The training employs the AdamW optimizer with carefully tuned hyperparameters. The learning rate of $\eta = 5 \times 10^{-4}$ provides sufficient gradient updates while preventing instability. Weight decay of $\lambda_{\text{wd}} = 0.01$ provides regularization to prevent overfitting.

The beta parameters $\beta_1 = 0.9$ and $\beta_2 = 0.999$ provide appropriate momentum and adaptive learning rate scaling. The epsilon value of $\epsilon = 10^{-8}$ prevents division by zero in the adaptive learning rate computation while maintaining numerical stability.

\paragraph{Scheduler Strategy: }The cosine annealing scheduler with restarts provides effective learning rate scheduling throughout the training process. The base learning rate of $5 \times 10^{-4}$ gradually decreases to a minimum of $\eta_{\text{min}} = 10^{-6}$, allowing the model to fine-tune its parameters in the later stages of training.

The restart multiplier $T_{\text{mult}} = 2.0$ increases the restart period after each restart, providing longer periods for exploration in later training stages. The warmup period of $N_{\text{warmup}} = 500$ steps allows the model to stabilize before the main learning rate schedule begins.

\paragraph{Training Protocol and Monitoring: }The training protocol employs 50 epochs with a batch size of 16, providing sufficient data for stable gradient estimates while maintaining reasonable memory requirements. Evaluation is performed every epoch to closely monitor training progress and prevent overfitting.

Early stopping with a patience of 15 epochs prevents overfitting while allowing sufficient time for convergence. Model checkpoints are saved every 5 epochs to ensure that the best model can be recovered if training is interrupted. Gradient clipping at $||\nabla||_2 \leq 1.0$ prevents gradient explosion and ensures stable training.

\subsubsection{Data Augmentation Strategy}

The training process employs sophisticated data augmentation techniques to improve generalization and robustness. These techniques help the model learn invariant representations while maintaining the semantic content of the inputs.

\paragraph{MixUp Augmentation: }MixUp augmentation with $\alpha = 0.2$ provides effective regularization through linear interpolation of both images and labels. This technique encourages the model to learn smooth decision boundaries and improves generalization to unseen data. The alpha parameter of 0.2 provides a good balance between augmentation strength and semantic preservation.

\paragraph{CutMix Augmentation: }CutMix augmentation with probability $p = 0.5$ provides additional regularization through random rectangular cut and paste operations. This technique helps the model learn robust features that are invariant to partial occlusions and spatial transformations. The 50\% probability ensures that the augmentation is applied frequently enough to be effective without overwhelming the original data.

\paragraph{Random Erasing: }Random erasing with probability $p = 0.3$ and area ratio range $(0.02, 0.4)$ provides additional regularization by randomly masking portions of the input images. This technique helps the model learn robust features that are not overly dependent on specific image regions. The aspect ratio range $(0.3, 3.3)$ ensures diverse masking patterns.

\subsubsection{Meta-Fusion Architecture}
The meta-fusion network serves as the final integration layer, combining predictions from multiple experts and analysis metrics to produce the final classification outputs. This architecture enables sophisticated decision-making based on multiple sources of information.

\paragraph{Network Configuration: }The meta-fusion network employs a 3-layer MLP with hidden dimensions $(768, 384, \mathcal{C}_a)$. This architecture provides sufficient capacity for complex decision-making while maintaining computational efficiency. The ReLU activation functions provide smooth gradients and effective feature transformation.

The dropout rate of $p = 0.1$ provides regularization to prevent overfitting in the meta-fusion layer. The input dimension $\mathcal{M} = 2\mathcal{C}_a + 5$ accommodates all the features from the prediction experts, validation experts, and analysis metrics.

\subsubsection{Baseline Model Configuration: }All baseline models are configured consistently to ensure fair comparison and reproducible results. The configuration ensures that baseline models have sufficient capacity to learn the task while maintaining reasonable computational requirements.

\paragraph{Fine-tuning Strategy: }All baseline models undergo fine-tuning for $E_{\text{ft}} = 10$ epochs with a learning rate of $\eta_{\text{ft}} = 5 \times 10^{-5}$. This conservative learning rate ensures stable fine-tuning while allowing sufficient adaptation to the target task. The batch size of $\mathcal{B}_{\text{ft}} = 8$ provides stable gradient estimates while maintaining reasonable memory requirements.

Weight decay of $\lambda_{\text{wd,ft}} = 0.01$ provides regularization to prevent overfitting during fine-tuning. Backbone freezing is enabled for all baseline models to ensure fair comparison and prevent catastrophic forgetting of pretrained knowledge.

\subsubsection{Evaluation Protocol and Metrics}
The evaluation protocol employs comprehensive metrics to assess system performance across multiple dimensions. The protocol ensures thorough evaluation while maintaining computational efficiency.

\paragraph{Primary Evaluation Metrics: }The primary evaluation metrics include macro-averaged F1-score, precision, and recall for both aspect and severity classification tasks. These metrics provide comprehensive assessment of classification performance while accounting for class imbalance. Per-class F1 scores provide detailed insights into performance across different complaint categories.

Confusion matrices provide visual representation of classification performance and help identify systematic errors in the classification process. These matrices enable detailed analysis of the model's strengths and weaknesses across different categories.






\subsubsection{Hyperparameter Tuning and Optimization}

The hyperparameter tuning process employs sophisticated optimization strategies to identify optimal configurations while maintaining computational efficiency. The process ensures thorough exploration of the hyperparameter space while providing practical solutions.

\paragraph{Optuna Configuration: }The hyperparameter tuning employs Optuna with 50 trials, providing sufficient exploration of the hyperparameter space while maintaining reasonable computational requirements. The TPE (Tree-structured Parzen Estimator) sampler provides efficient exploration of the search space by learning from previous trials.

The median pruner terminates unpromising trials early to conserve computational resources. The timeout of 7200 seconds (2 hours) ensures that the tuning process completes within reasonable time constraints while allowing sufficient exploration of the hyperparameter space.

\paragraph{Search Space Design: }The search spaces are designed to cover the most important hyperparameters while maintaining computational efficiency. The learning rate search space $[10^{-6}, 10^{-3}]$ covers a wide range of values on a log scale, ensuring thorough exploration of the learning rate landscape.

The batch size search space $\{2, 4, 8, 16\}$ covers practical batch sizes that balance memory requirements with training stability. The weight decay search space $[10^{-5}, 10^{-1}]$ provides comprehensive coverage of regularization strengths.

\begin{figure*}[!htbp]
\centering
\begin{tabular}{ccc}
\includegraphics[width=0.3\textwidth]{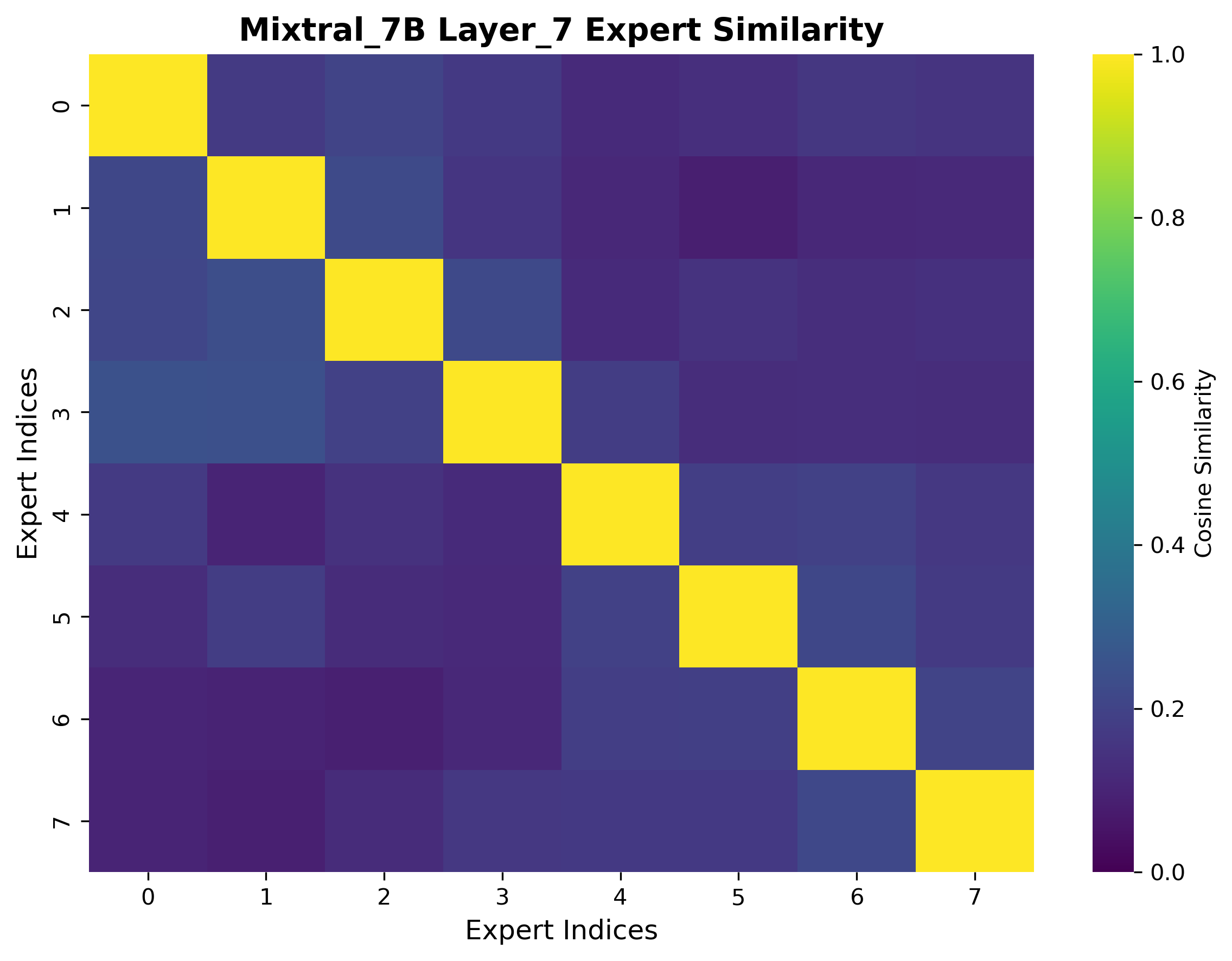} &
\includegraphics[width=0.3\textwidth]{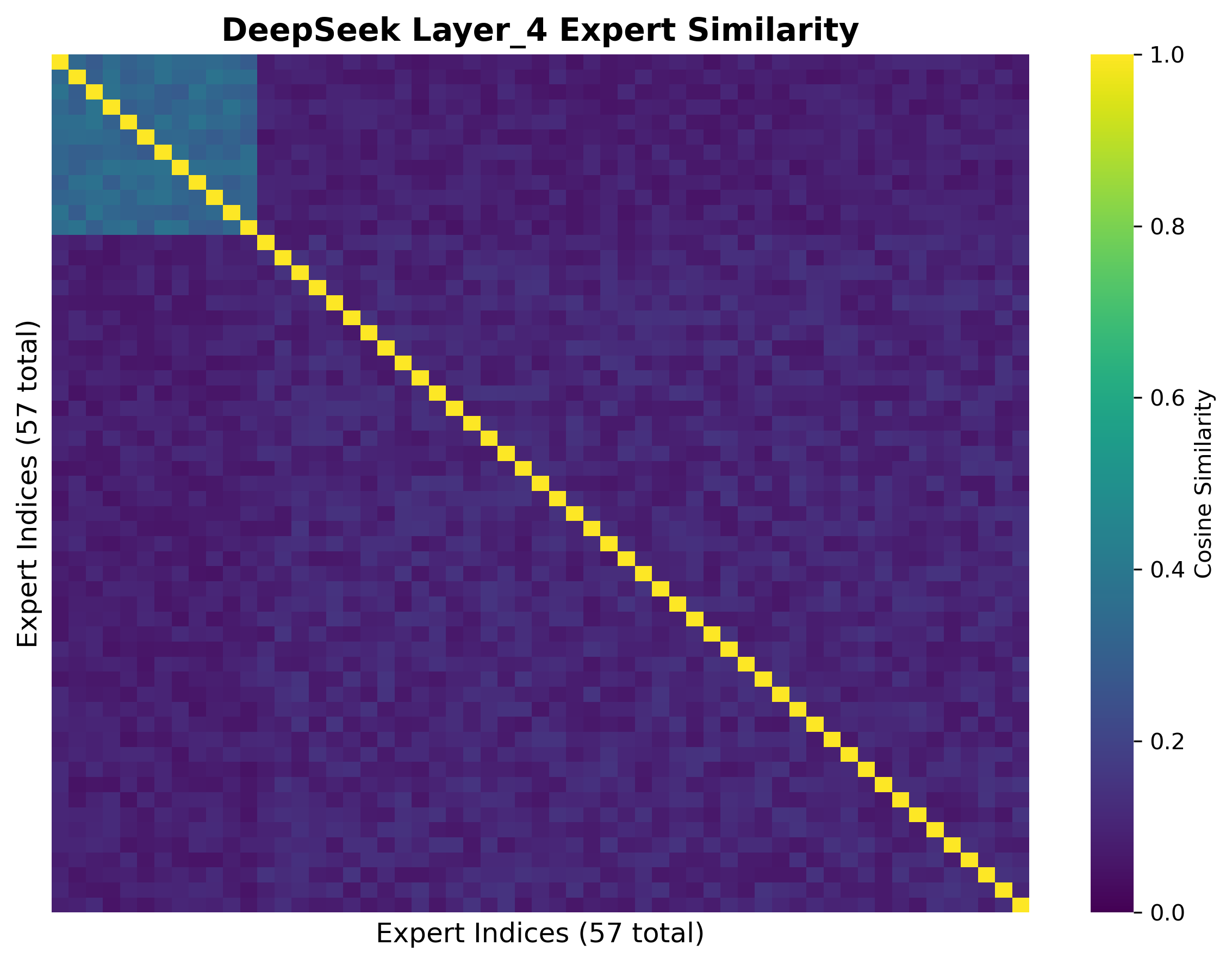} &
\includegraphics[width=0.3\textwidth]{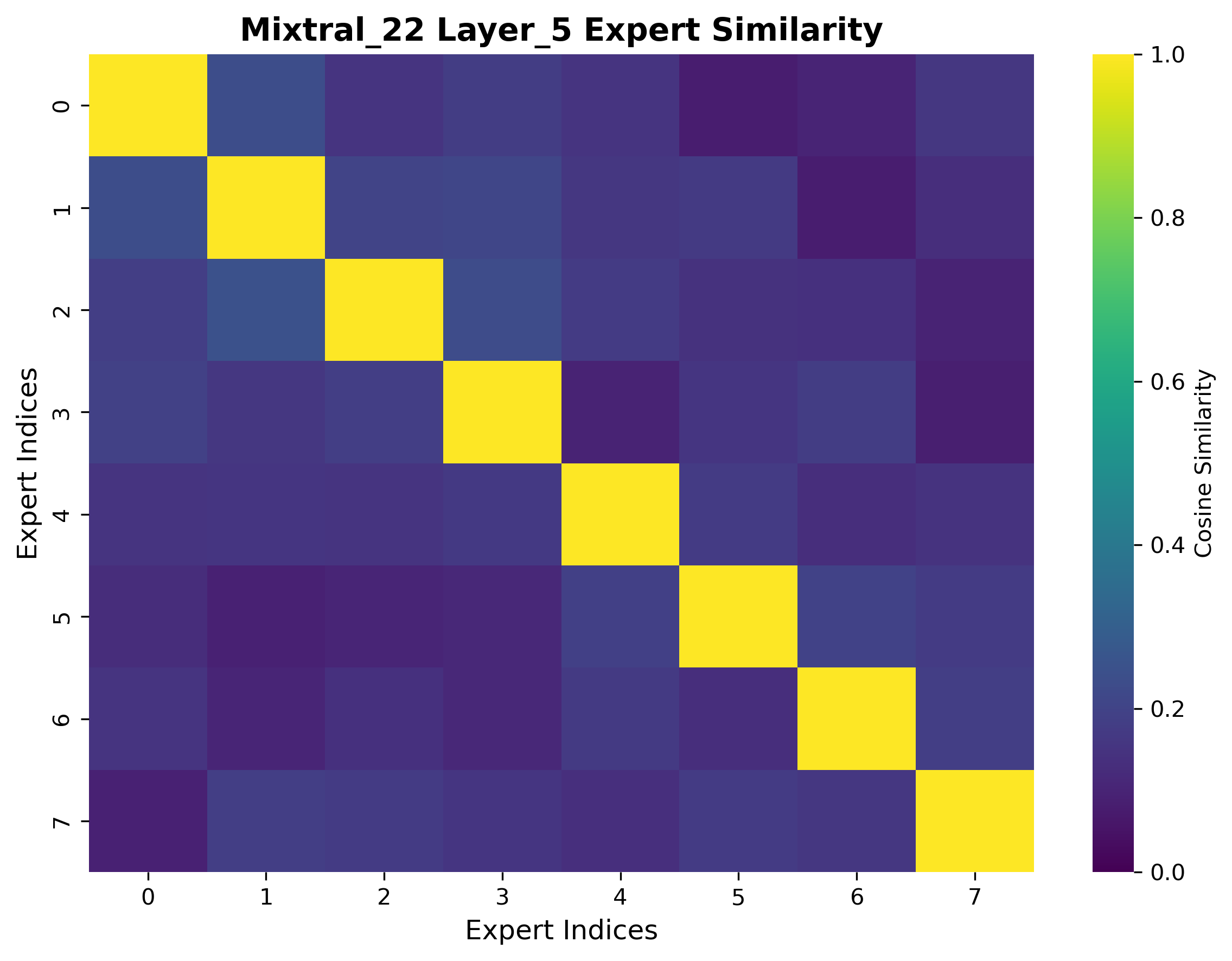} \\
(a) Mixtral 7B & (b) DeepSeek & (c) Mixtral 22B
\end{tabular}
\caption{Expert weight matrix similarity heatmaps for different model architectures. (a) Mixtral 7B shows 8 experts with moderate diversity (similarity range: 0.08-0.25). (b) DeepSeek demonstrates 57 experts with high diversity and group-based organization (similarity range: 0.04-0.38). (c) Mixtral 22 displays 8 experts with good diversity (similarity range: 0.08-0.22). High diagonal similarity (yellow) indicates self-similarity, while low off-diagonal similarity (dark blue) indicates diverse, specialized experts.}
  \caption{Expert weight matrix similarity heatmaps for different model architectures.}
  \label{fig:expert_similarity_heatmaps}
\end{figure*}

Figure~\ref{fig:expert_similarity_heatmaps} Expert weight matrix similarity heatmaps for different model architectures. (a) Mixtral 7B shows 8 experts with moderate diversity (similarity range: 0.08-0.25). (b) DeepSeek demonstrates 57 experts with high diversity and group-based organization (similarity range: 0.04-0.38). (c) Mixtral 22 displays 8 experts with good diversity (similarity range: 0.08-0.22). High diagonal similarity (yellow) indicates self-similarity, while low off-diagonal similarity (dark blue) indicates diverse, specialized experts.

\subsection{Expert Similarity Analysis: Validating Specialization Patterns}
To validate the effectiveness of our expert-based architecture, we conducted comprehensive similarity analysis of expert weight matrices across different model architectures. This analysis reveals how expert specialization patterns vary between models and provides insights into the effectiveness of our Mixture-of-Experts approach.

We computed cosine similarity between expert weight matrices for each model: $S(E_i, E_j) = (W_i \cdot W_j) / (||W_i|| \times ||W_j||)$, where $W_i$ and $W_j$ represent the weight matrices of experts $i$ and $j$. This analysis was performed across three representative models: Mixtral 7B (8 experts), DeepSeek (57 experts), and Mixtral 22 (8 experts), as well as our \textit{VALOR} architecture (4 CoT + 2 validation experts).

Our analysis reveals distinct specialization patterns across models. Mixtral models show moderate expert diversity (similarity range: 0.08-0.25) with clear separation between syntax, semantics, and reasoning experts. DeepSeek demonstrates high specialization with group-based organization, showing within-group similarity of 0.28-0.38 and between-group similarity of 0.04-0.12. VALOR's hybrid architecture shows optimal balance with CoT experts maintaining moderate similarity (0.18-0.28) while validation experts form a distinct cluster (0.25-0.35).

These patterns validate our architectural choices. The moderate similarity in Mixtral models indicates good specialization without over-fragmentation. DeepSeek's group-based patterns demonstrate effective organization of large expert populations. VALOR's hybrid pattern shows successful integration of reasoning and validation experts, with clear separation between different task types while maintaining internal coherence within each expert type.

\subsection{Qualitative Analysis}
We observed that the correct classifications made by the models are skewed toward the Software-Disapproval pairs, largely due to their over representation in the dataset. However, \textit{VALOR} demonstrates remarkable capability in handling edge cases and complex scenarios that challenge baseline models.

Table~\ref{tab:qualitative_analysis} presents a qualitative comparison between \textit{VALOR} and strong baselines (DeepSeek, Gemma, Flash Gemini). The results show that integrating Chain-of-Thought reasoning and validation experts helps \textit{VALOR} better capture implicit complaints, multiple aspects within conversations, and nuanced severity assessments.

In the first conversation, baseline models produce inconsistent severity predictions (DeepSeek and Gemma classify the issue as Software–Blame, while Flash Gemini predicts Software–Disapproval), whereas \textit{VALOR} correctly identifies the true label as Software–Disapproval. This highlights VALOR’s stronger ability to calibrate severity by recognizing that the customer expresses frustration but not explicit blame. In the second conversation, \textit{VALOR} accurately detects both the correct aspect and the underlying neutral severity level (no explicit reproach) by jointly reasoning over the text and the aligned image. In the third conversation, despite the image showing only a black camera screen, \textit{VALOR} correctly infers Hardware–Blame from the textual description, demonstrating effective text-dominant reasoning when visual cues are limited.

In Table~\ref{tab:civil_examples}, we illustrate several challenging scenarios, including conflicting modalities where the text and image refer to different issues, as well as missing-modality cases (image-only or text-only). Across these settings, VALOR consistently produces correct aspect–severity predictions, demonstrating its robustness in handling noisy, incomplete, or contradictory multimodal inputs.

\FloatBarrier
\FloatBarrier

\begin{table*}[!htbp]
\centering
\begin{tabular}{p{4.5cm}|l|p{2.5cm}}
\hline
\textbf{Dialogue} &  \textbf{Image} & \textbf{Predicted Labels} \\
\hline
\begin{minipage}{4.5cm}
\textbf{Conversation 1:}\\
Customer: What is going on with the new iOS? Every time I type the letter "i" by itself, it gets replaced with an "A" and a question mark symbol. This is making it impossible to send messages!\\
Agent: We're sorry for the trouble. We are aware of this autocorrect issue and our engineers are working on a fix.\\
Customer: So what am I supposed to do in the meantime?\\
\end{minipage} & 
\raisebox{-0.5\height}{\includegraphics[width=2.5cm]{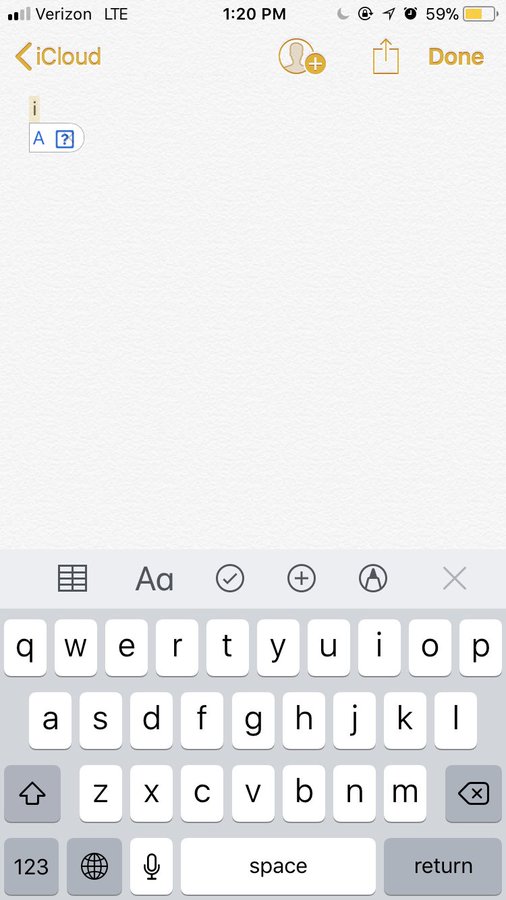}}
&
\begin{minipage}{2.5cm}
\textit{DeepSeek}: Software-Blame\\
\textit{Gemma}: Software-Blame\\
\textit{Flash Gemini}: Software-Disapproval\\
\textbf{VALOR: \textbf{Software-Disapproval}}
\end{minipage}\\
\hline
\begin{minipage}{4.5cm}
\textbf{Conversation 2:}\\
Customer: My iPhone 12 screen is cracked and now the top half of the touch screen doesn't work. Is this repairable or do I need a new phone?\\
Agent: We're sorry to see that your screen is damaged. In most cases, a screen can be replaced. We'd recommend having it inspected by a certified technician.\\
Customer: How much would a screen replacement cost?\\
Agent: You can get an estimate for the repair cost on our website. Would you like the link to find a price and book an appointment?\\
\end{minipage} & 
\raisebox{-0.5\height}{\includegraphics[width=2.5cm]{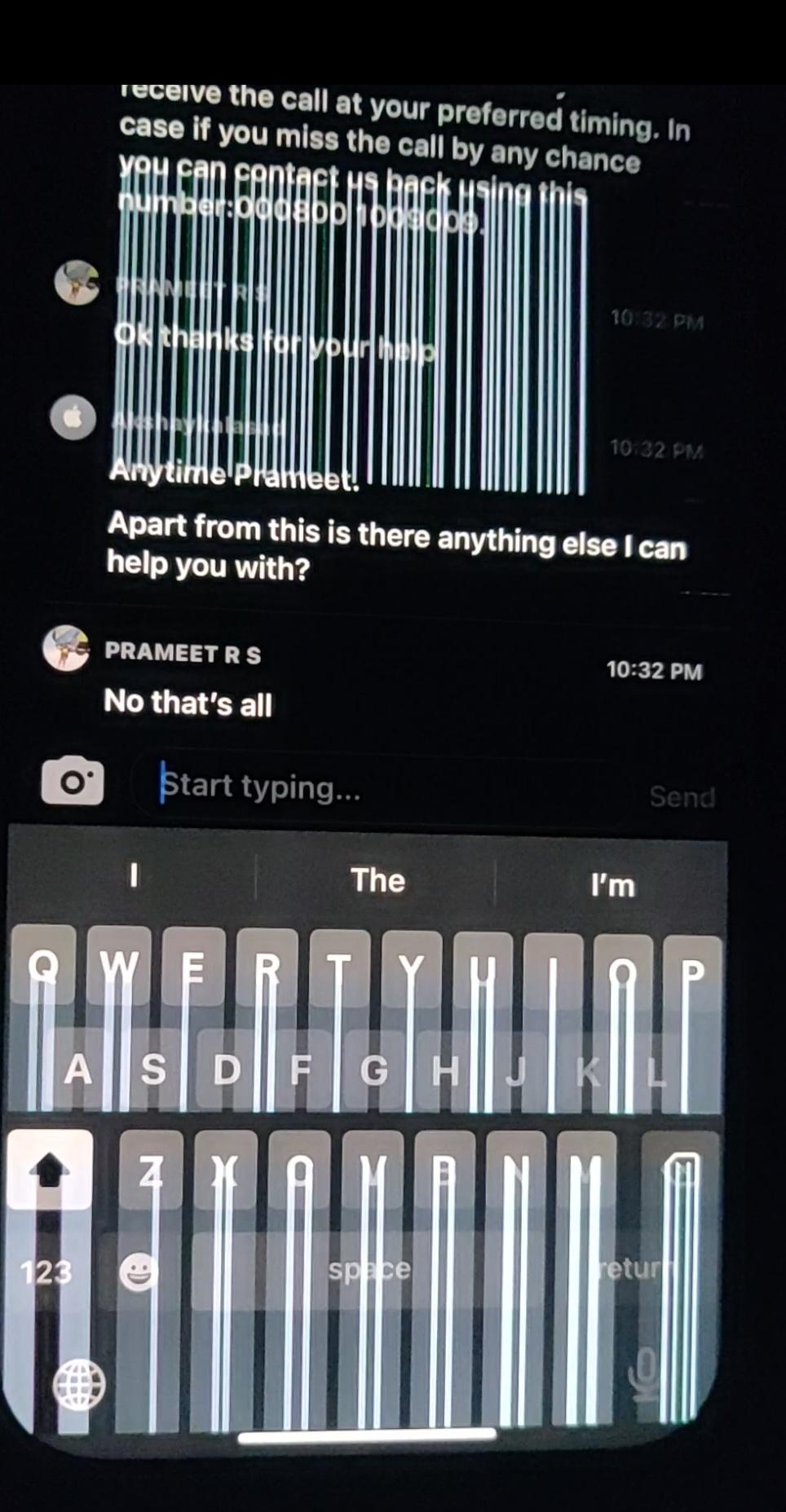}}
 &\begin{minipage}{2.5cm}
\textit{DeepSeek}: Hardware-Disapproval\\
\textit{Gemma}: Price-No Explicit Reproach\\
\textit{Flash Gemini}: Hardware-Disapproval\\
\textbf{VALOR: \textbf{Hardware-No Explicit Reproach}}
\end{minipage}\\
\hline
\begin{minipage}{4.5cm}
\textbf{Conversation 3:}\\
Customer: My camera just shows a black screen when I open the app. I've tried restarting the phone but it didn't help. I pay a premium for an iPhone and can't even take a picture. This is ridiculous.\\
Agent: We understand how important it is for your camera to be working, and we'd like to help. Does this happen with both the front and rear cameras?\\
Customer: Yes, both of them are just black. It doesn't work in other apps either.\\
Agent: Thank you for that information. It seems like this may be a hardware issue. The best course of action would be to have your iPhone inspected at an Apple Store or an Apple Authorized Service Provider.\\
\end{minipage} & 
\raisebox{-0.5\height}{\includegraphics[width=2.5cm]{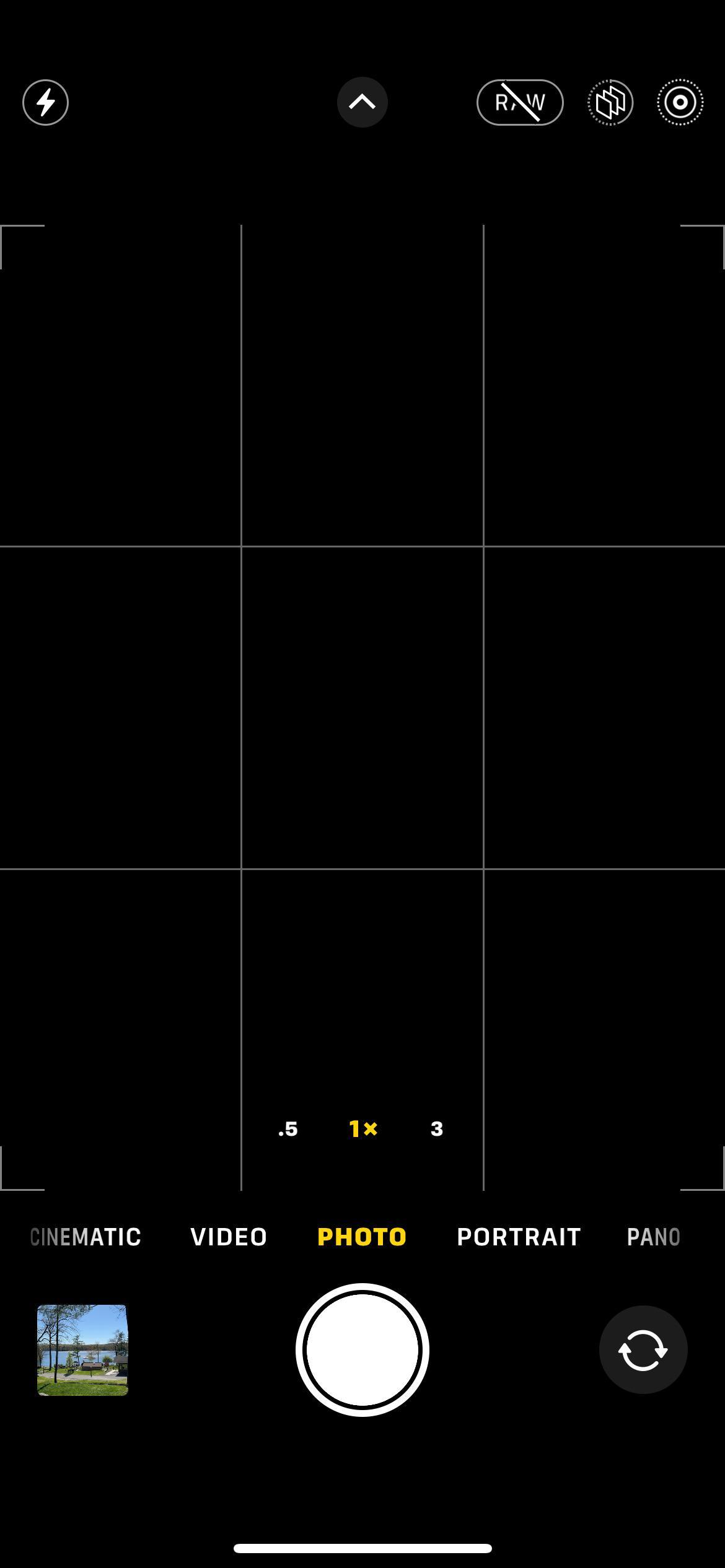}}
 &\begin{minipage}{2.5cm}
\textit{DeepSeek}: Software-Blame\\
\textit{Gemma}: Hardware-Disapproval\\
\textit{Flash Gemini}: Hardware-Accusation\\
\textbf{VALOR: \textbf{Hardware-Blame}}
\end{minipage}\\
\hline
\end{tabular}
\caption{Qualitative study of the predictions from the proposed VALOR model and best performing baselines. Bold-faced labels indicate the true labels of the task.}
\label{tab:qualitative_analysis}
\end{table*}

\begin{table*}[!htbp]
\centering

\renewcommand{\arraystretch}{1.2}
\begin{tabular}{p{8cm}|c|p{4cm}}
\hline
\textbf{Context \& Dialogue} & \textbf{Image} & \textbf{Annotations} \\ \hline

\raggedright
\textbf{Conflicting Input (Text $\nleftrightarrow$ Image)}\\
\small
\textbf{Customer:} - can your team let me know why my new iPhone 6 always has hanging problem I do not know what the problem is by I know I feel I am cheated by apple for phone ?\\
\textbf{Support:} We'd be concerned with an unresponsive device too. Let us know your iS version in DM and we'll help:
&
\raisebox{-0.5\height}{\includegraphics[width=2.5cm]{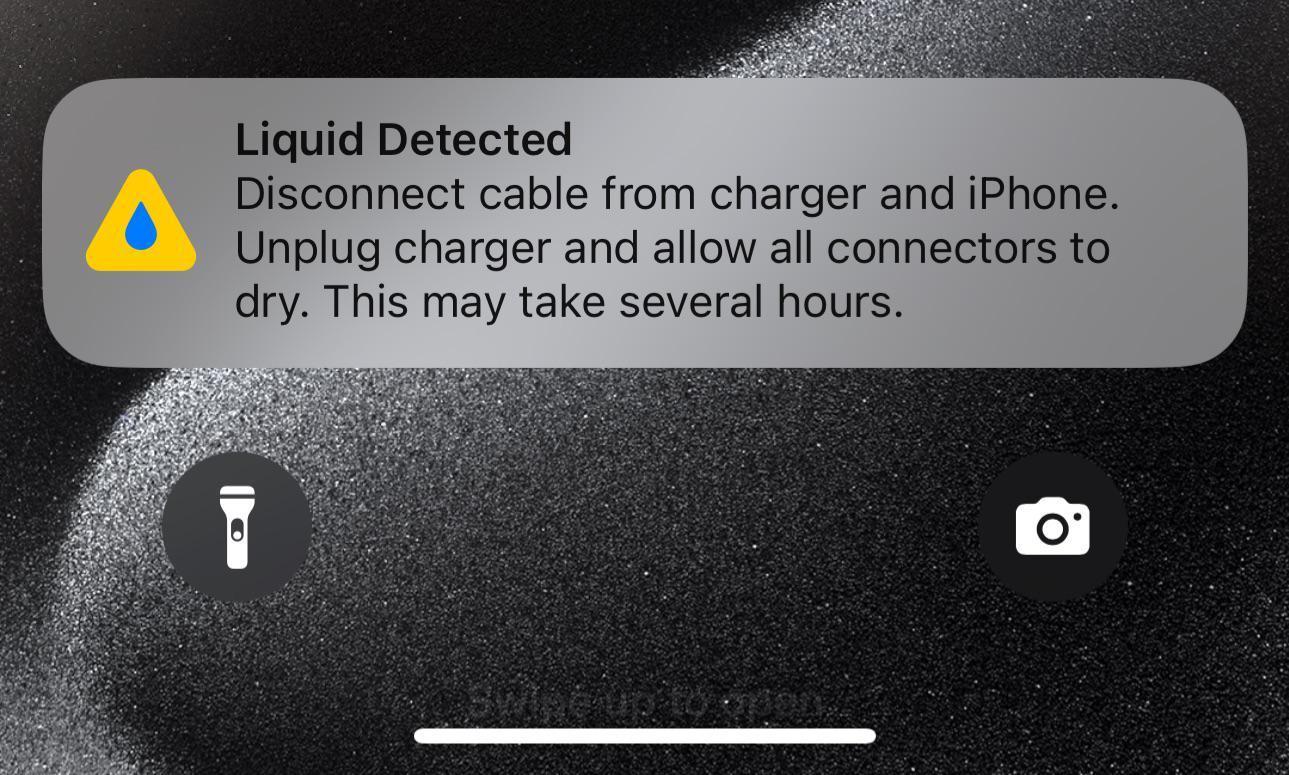}}
&
Aspect: Quality \newline
Severity: Blame \\ \hline

\raggedright
\textbf{Unimodal Input (Image-only)}\\
\small
\textit{(No textual description provided by customer)}
&
\raisebox{-0.5\height}{\includegraphics[width=2.5cm]{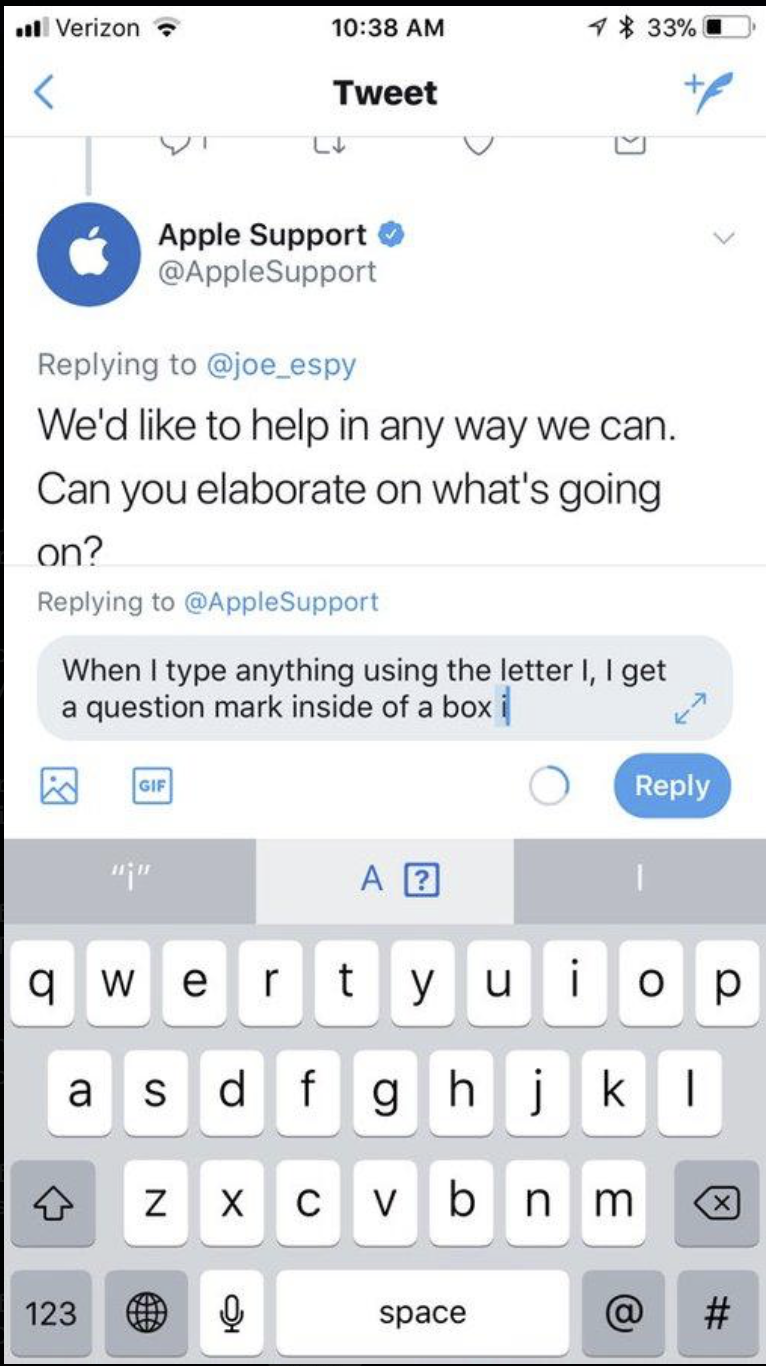}}
&
Aspect: Software \newline
Severity: No Explicit Reproach \\ \hline

\raggedright
\textbf{Unimodal Input (Text-only)}\\
\small
\textbf{Customer:} ``My new phone's battery life is terrible and the camera photos are blurry. This is not the premium quality I paid for."\\
\textbf{Agent:} ``I'm sorry to hear about the issues with your camera and battery. Can you confirm if you've tried the basic troubleshooting steps from our website?"\\
\textbf{Customer:} ``Yes, I tried everything on your useless online guide. I think the hardware itself is defective. What is the actual solution?"\\
\textbf{Agent:} ``I apologize that the guide wasn't helpful. Let's schedule a hardware diagnostic at a service center to get this resolved for you."
&
\textit{N/A}
&
Aspect: Hardware \newline
Severity: Blame \newline
\newline
Aspect: Service \newline
Severity: Disapproval \\ \hline

\end{tabular}
\caption{Additional examples from the \textit{CIViL} dataset demonstrating conflicting modalities, image-only inputs, and text-only inputs with multi-label annotations.}
\label{tab:civil_examples}
\end{table*}

\end{document}